\documentclass[pdflatex,sn-mathphys-num]{sn-jnl}                                        
\usepackage{graphicx}
\usepackage{multirow}
\usepackage{amsmath,amssymb,amsfonts}
\usepackage{amsthm}
\usepackage{float}      
\usepackage{rsfso}
\usepackage[title]{appendix}
\usepackage{xcolor}
\usepackage{textcomp}
\usepackage{manyfoot}
\usepackage{booktabs}
\usepackage{algorithm}
\usepackage{algorithmicx}
\usepackage{algpseudocode}
\usepackage{listings}                                           \usepackage{stfloats}                                                               \usepackage{comment}                                             
\usepackage{placeins}   
\usepackage{bookmark}
\bookmarksetup{startatroot}
\theoremstyle{thmstyleone}

\DeclareUnicodeCharacter{2061}{}
\theoremstyle{thmstyletwo}

\theoremstyle{thmstylethree}

\begin{document}

\title[SACA Clustering Algorithm]{SACA: Selective Attention-Based Clustering Algorithm}

\author[1,2,3,4]{\fnm{Meysam} \sur{Shirdel Bilehsavar}}\email{meysam@email.sc.edu}
\author[5]{\fnm{Razieh} \sur{Ghaedi}}\email{GhaediR@cardiff.ac.uk}
\author[1,2,3,4]{\fnm{Christian} \sur{O'Reilly}}\email{christian.oreilly@sc.edu} 

\affil[1]{\orgdiv{Department of Computer Science}, \orgname{University of South Carolina}, \city{Columbia}, \country{USA}}
\affil[2]{\orgdiv{Artificial Intelligence Institute}, \orgname{University of South Carolina}, \city{Columbia}, \country{USA}}
\affil[3]{\orgdiv{Carolina Autism and Neurodevelopment Research Center}, \orgname{University of South Carolina}, \city{Columbia}, \country{USA}}
\affil[4]{\orgdiv{Institute for Mind and Brain}, \orgname{University of South Carolina}, \city{Columbia}, \country{USA}}
\affil[5]{\orgdiv{Cardiff Business School}, \orgname{Cardiff University}, \city{Cardiff}, \country{UK}}

\vspace{-4pt}

\abstract{Clustering algorithms are fundamental tools across many fields, with density-based methods offering particular advantages in identifying arbitrarily shaped clusters and handling noise. However, their effectiveness is often limited by the requirement of critical parameter tuning by users, which typically requires significant domain expertise. This paper introduces a novel density-based clustering algorithm loosely inspired by the concept of selective attention, designed to minimize reliance on parameter tuning for most applications. The proposed method computes an adaptive threshold to exclude sparsely distributed points and outliers, constructs an initial cluster framework, and subsequently reintegrates the filtered points to refine the final results. Extensive experiments on diverse benchmark datasets demonstrate the robustness, accuracy, and ease of use of the proposed approach, establishing it as a powerful alternative to conventional density-based clustering techniques.}

\keywords{Clustering, Density-Based, Selective Attention, Unsupervised Learning, Data Mining.}


\maketitle
\section{Introduction}
\label{introduction}

Clustering is a core data mining task that aims to identify groups of related items within a dataset while segregating dissimilar things into distinct groups \cite{weng2021hdbscan, Campello2020}. It has an important role in understanding data features, relationship between samples, and the overall structure of information in a dataset \cite{Zhuang2020}. As an unsupervised technique, clustering does not require training or labeled data \cite{Jain1999,martos2023clustering,oyewole2023data}, which makes it attractive in situations where distinct categories are suspected but unknown. In recent years, algorithmic advancement in this field has attracted considerable interest from both academia and industry \cite{Cai2020b}. Although clustering algorithms are widely used in tasks such as pattern analysis, grouping, decision-making, document retrieval, image segmentation, and data mining, significant challenges remain \cite{Bhattacharjee2021}. For instance, the result of widely used algorithms such as \textit{k}-Means often depends on initial conditions (e.g., initial cluster centroids). The effect of such initial conditions on the solution can be mitigated by using multiple runs with different initializations and comparing outcomes. However, these algorithms are not guaranteed to find the globally optimal solution. Moreover, the iterative computations performed by such algorithms are laborious, computationally expensive, and prone to uncertainty \cite{Zou2020}. The presence of noise further complicates the problem, reducing the reliability of results while increasing processing and storage costs \cite{Cai2022}. 

Clustering techniques can be categorized into five types: partition-based, hierarchical, density-based, model-based, and grid-based \cite{Soni2012}. Among these, density-based clustering algorithms are particularly advantageous because they eliminate the need to manually specify the number of clusters and are effective in handling noisy data \cite{ester1996density}. Density-Based Spatial Clustering of Applications with Noise (DBSCAN) \cite{Ester1996}, is a prominent clustering technique that provided a foundation for density-based clustering. Its ability to discover clusters of arbitrary shapes, exclude noise points, and operate without a predefined number of clusters makes it especially attractive. However, DBSCAN faces challenges, such as difficulty in detecting clusters with heterogeneous densities \cite{Wang2020} and in distinguishing adjacent clusters \cite{Tran2013}. Further, it is computationally expensive on large datasets due to its iterative nature \cite{Ester1996}. Its performance also depends heavily on the careful selection of two parameters: the maximum radius (Eps) and the minimum number of neighbors (minPts) \cite{Li2020}. Automatically determining these parameters is challenging, particularly for extensive datasets.

This paper tackles a key limitation of conventional density-based clustering algorithms, their strong dependence on fine-tuned input parameters (two in the case of DBSCAN). We introduce a novel algorithm, the Selective Attention-Based Clustering Algorithm (SACA), loosely inspired by the concept of selective attention. SACA substantially reduces the reliance on user-defined parameters by computing an adaptive threshold derived from intrinsic dataset characteristics to temporarily exclude sparse or low-density points. A preliminary cluster structure is then established, after which the excluded points are systematically reintegrated to finalize the clustering outcome. In most practical applications, SACA operates without any need for parameter tuning. In rare scenarios where clusters exhibit extreme overlap, a single, intuitive integer parameter, the Attention Selectivity Coefficient ($C$), provides a simple means to adjust sensitivity. Increasing $C$ enhances the algorithm’s ability to identify localized structures, providing control over the clustering resolution in intricate or multi-level datasets. This mechanism offers a clear and interpretable guideline for parameter adjustment, eliminating the need for extensive trial and error or specialized domain knowledge. Altogether, SACA offers three major contributions to the field:

\begin{enumerate}

    \item \textbf{Inspired by selective attention mechanism:} SACA builds on the concept of selective attention (i.e., the ability of ignoring irrelevant information) by focusing on high-density core structures to distinguish structural patterns from noise.

    \item \textbf{Intuitive parameter control:} When needed, the Attention Selectivity Coefficient ($C$) provides straightforward adjustment for complex scenarios, enabling multilevel pattern discovery from global to local structures.

    \item \textbf{Superior performance with minimal tuning:} SACA performs well in most cases with no or little parameter tuning. Comprehensive validation across 16 benchmark datasets demonstrates consistent superiority over DBSCAN, HDBSCAN, and OPTICS across six metrics, requiring substantially fewer adjustments.
    
\end{enumerate}

The rest of this paper is organized as follows. \textit{Section 2} presents prominent density-based clustering techniques and examines their limitations.  \textit{Section 3} outlines the concept of selective visual attention and discuss how it can offer inspiration for improving clustering. \textit{Section 4} describes the proposed method and provides a pseudocode for SACA. \textit{Section 5} describes the datasets, evaluation metrics, and experimental setup. \textit{Section 6} reports and analyses the results including multilevel pattern recognition. \textit{Section 7} discusses limitations and directions for future research. Finally, \textit{Section 8} concludes the paper.

\section{Related Work }
\label{sec:MILP-models}

Unlike centroid-based methods like k-Means, which assume spherical clusters \cite{jain2010data}, density-based clustering algorithms discover arbitrarily shaped clusters and handle noise by grouping points based on local density. However, many of these methods critically depend on user-defined parameters requiring domain expertise or extensive tuning. This section reviews prominent density-based methods, organizes them by how they handle parameters, and highlights the remaining gap that motivates SACA.

\subsection{Classical Density-Based Methods}

The seminal DBSCAN algorithm identifies clusters by expanding regions with high point density, defined by two parameters: the neighborhood radius (\(\varepsilon\)) and the minimum number of points (\textit{MinPts}). Points outside dense regions are labeled as noise. DBSCAN excels at detecting arbitrarily shaped clusters but struggles with datasets containing clusters of varying density, as a single global \(\varepsilon\) is often inadequate. Its performance is highly sensitive to parameter choices, necessitating trial-and-error tuning, and its computational cost increases with large datasets due to repeated distance calculations \cite{ester1996density}. OPTICS (Ordering Points To Identify the Clustering Structure) \cite{ankerst1999optics} mitigates sensitivity to a fixed \(\varepsilon\) by producing a hierarchical ordering using core and reachability distances. As with DBSCAN, it requires \(\varepsilon\) (a maximum neighborhood radius) and \textit{MinPts}, but is less sensitive to \(\varepsilon\) and better captures clusters across varying density levels. However, OPTICS demands additional memory for nearest-neighbor queries and can struggle when adjacent clusters have similar local densities. DENCLUE (DENsity-based CLUstEring) \cite{10.5555/3000292.3000302,hinneburg1998efficient} models point density as a sum of influence functions (e.g., Gaussian kernels) and finds density attractors via a hill-climbing procedure. It requires two parameters: \(\sigma\) (influence range) and \(\xi\) (density threshold). While effective in high dimensions and robust to noise, DENCLUE is sensitive to \(\sigma\) and \(\xi\), and hill-climbing can converge to local maxima.

\subsection{Adaptive Variants }

A second line of work retains DBSCAN-like principles but makes parameter selection more adaptive. VDBSCAN (Varied Density-Based Spatial Clustering of Applications with Noise) \cite{liu2007vdbscan} selects multiple \(\varepsilon\) values based on \(k\)-distance plots, partitions the dataset by density levels, and applies DBSCAN-like clustering for each partition improving adaptability to heterogeneous densities. However, it still depends on precise \(\varepsilon\) and \textit{MinPts} settings and can struggle with non-convex clusters. DVBSCAN (Density Variation Based Spatial Clustering of Applications with Noise) \cite{ram2010density} introduces density-variance controls (Cluster Density Variance and Cluster Similarity Index) to allow within cluster density variation while separating distinct density profiles, but requires four parameters (\(\varepsilon\), \textit{MinPts}, \(\alpha\), \(\lambda\)), increasing tuning complexity. ST-DBSCAN \cite{birant2007st} extends DBSCAN to spatial–temporal data with two distance parameters (\(\varepsilon_1\) spatial, \(\varepsilon_2\) non-spatial), \textit{MinPts}, and a threshold \(\Delta \epsilon\) to avoid merging adjacent clusters. It incorporates a density factor to address varied density and temporal dependence. GF-DBSCAN (Global Filtering DBSCAN) \cite{tsai2009gf} reduces search space by partitioning data into grid cells based on \(\varepsilon\), which can improve scalability but may sacrifice accuracy. AutoEpsDBSCAN \cite{gaonkar2013autoepsdbscan} automates \(\varepsilon\) selection using \(k\)-nearest neighbor distances and elbow detection in \(k\)-distance plots. It removes manual \(\varepsilon\) tuning in favorable cases but still requires choosing \(k\). Further, the performance of this algorithm drops when clusters transition gradually into noise (i.e., lack sharp density boundaries).

\subsection{Parameter-Free/Reduced Approaches}
A third direction reduces or eliminates user-defined settings. DBCLASD (Distribution-Based Clustering of Large Spatial Databases) \cite{xu1998distribution} grows clusters by testing nearest neighbor distances against a uniform-distribution model, removing explicit \(\varepsilon\)/\textit{MinPts} selection. However, its uniform-density assumption can misclassify points in non-uniform regions. More recent approaches further reduce sensitivity by leveraging neighborhood structure. NNVDC (Nearest Neighbor Variable Density-based Clustering) \cite{hou2024density} uses a single parameter that defines the number of nearest neighbors \(k\) to form clusters via mutual \(k\)-NN relationships. While simpler than multi-parameter methods, performance remains sensitive to \(k\), potentially causing over- or under-clustering. PDCSN (Partition Density Clustering with Self-Adaptive Neighborhoods) \cite{xing2023pdcsn} adapts neighborhood size \(\lambda\) from natural nearest-neighbor density distributions and forms clusters from mutual \(\lambda\)-NN relationships, effectively distinguishing adjacent clusters with similar or varying densities. DPC-MDNN (Density Peak Clustering with Manifold Distance and Natural Nearest Neighbors) \cite{wang2025improved} improves detection by using manifold distances and natural neighbors to reflect intrinsic geometry, though it can be computationally demanding and implicitly sensitive to neighborhood scale.

\subsection{Summary}

The algorithms reviewed in this section are listed in Table \ref{tab:rw-compare}, along with the key idea they implement and the number of parameters they require. Collectively, these methods show clear progress toward adaptive and parameter-efficient density-based clustering. Nevertheless, most approaches retain some degree of parameter sensitivity (e.g., \(\varepsilon\), \textit{MinPts}, \(k\), \(\lambda\)) or rely on assumptions that may falter with heterogeneous densities, overlapping clusters, or multi-scale structure. In practice, this leads to trial-and-error tuning, dependence on expert knowledge, and reduced accessibility for non-experts. Building on these trends, SACA takes inspiration from the concept of selective attention to derives a global threshold directly from intrinsic dataset statistics and forms robust core structures, temporarily filtering sparse points. In Table \ref{tab:rw-compare}, SACA is listed as having 1+ parameters because it depends on $C$ and its behavior can also be modified by a boolean parameter \textit{use\_center} (described later). However, as we will show in our results, these parameters are intuitive, easy to use, seldom need to be set to values other than their default values, and, when they are needed, they can typically be fine-tuned with fewer trial-and-error steps than alternative algorithms. 

\begin{table}[h]
\centering
\footnotesize
\setlength{\tabcolsep}{3.5pt}
\caption{\normalsize Comparison of representative density-based clustering algorithms.}
\label{tab:rw-compare}
\begin{tabular}{llcc}
\toprule
\textbf{\small Method} & \textbf{\small Key Idea} & \textbf{\small Params}  \\
\midrule
DBSCAN \cite{ester1996density}        & Density reachability ($\varepsilon$, MinPts) & 2   \\
OPTICS \cite{ankerst1999optics}       & Reachability ordering (multi-density)        & 2   \\
DENCLUE \cite{hinneburg1998efficient} & Kernel density attractors                    & 2    \\
VDBSCAN \cite{liu2007vdbscan}         & Multiple $\varepsilon$ for varied densities  & 2+   \\
DVBSCAN \cite{ram2010density}         & Density-variance control                     & 4   \\
DBCLASD \cite{xu1998distribution}     & Distribution-based cluster growing           & 0   \\
AutoEpsDBSCAN \cite{gaonkar2013autoepsdbscan} & $k$-dist elbow for $\varepsilon$       & 1+   \\
NNVDC \cite{hou2024density}           & Mutual $k$-NN connectivity                   & 1  \\
PDCSN \cite{xing2023pdcsn}            & Self-adaptive radius ($\lambda$)             & 1   \\
DPC-MDNN \cite{wang2025improved}      & Density peaks + manifold NN                  & 0   \\
\midrule
\textbf{SACA (proposed)}               & \textbf{Data-derived threshold + reintegration} & \textbf{1+}  \\
\bottomrule
\end{tabular}
\end{table}

\FloatBarrier

\section{Visual Selective Attention}\label{subsec1}

Although we do not claim to have integrated formal elements from a cognitive model of visual selective attention into SACA, we drew inspiration from the qualitative features of this cognitive process. More specifically, we attempted to reproduce the concept of selective attention, which filters out irrelevant information while focusing on important data. Broadly, visual selective attention regulates the processing of retinal input based on relevance and selects particular representations to enter perceptual awareness \cite{Chelazzi2013}. This mechanism enables the selective focus on information pertinent to the current context while disregarding competing, less relevant stimuli \cite{Kotary1995}.

\begin{figure}[htbp]
\centering
\includegraphics[width=0.50\textwidth]{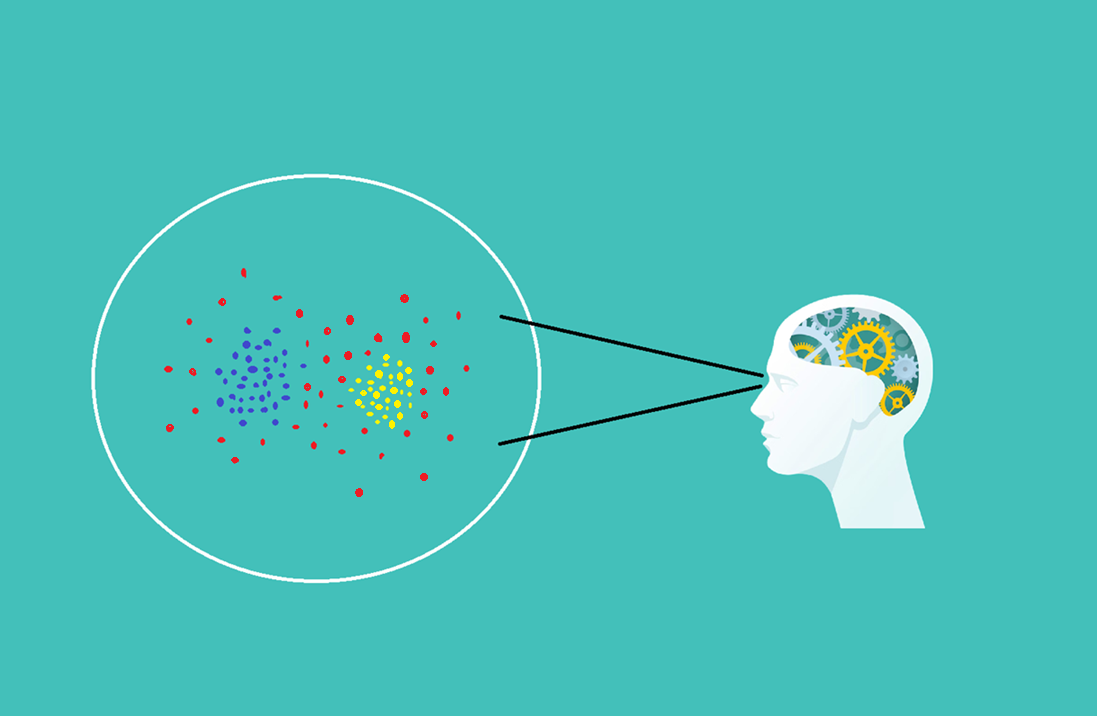}
\caption{Simplified depiction of visual selective attention. The brain attempts to find relevant patterns of information, emphasizing the importance of structured information (blue and yellow clusters) over unstructured noise (red dots).}
\label{fig: Figure 1}
\end{figure}

\FloatBarrier

The application of this idea to clustering is schematically represented in Figure~\ref{fig: Figure 1}. It relies on the assumption that, to make sense of complex visual scenes, humans naturally ignore irrelevant elements (i.e., data points, in the context of clustering) to selectively attend to information assumed to be important by virtue of co ntaining structurally identifiable patterns. The brain is inherently biased toward creating significant patterns from noisy input by emphasizing relevant signals and suppressing irrelevant ones. We observe this bias in our natural tendency to see meaningful forms in clouds or inkblots, as exploited, for example, by the Rorschach test \cite{weiner2013rorschach}. This tendency of attributing meaningful interpretation to nebulous, or noisy sensory input is known as pareidolia and has a long history \cite{kahlbaum1866sinnesdelirien}. It is also captured by the concept of predictive-coding \cite{rao1999predictive}, and the idea that our perceptions are not a raw representation of the information reaching the retina, but a fusion of the incoming data and our internal expectation about likely incoming data. At the root of this mechanism is the ability of our brain to emphasize the importance of some data and ignore others. 

By analogy, SACA attempts to mimic this functionality by retaining data points that contribute to the core structure of clusters while filtering out noisy or irrelevant points. In the next subsection, we translate this intuition into a concrete, dataset-derived pruning rule.

\section{Methods}
To address the limitations of existing algorithms, this paper proposes the Selective Attention-Based Clustering Algorithm (SACA). This clustering approach identifies clusters based on core structures separated from each other using a neighborhood threshold automatically calculated from the dataset. In the previous section, we introduced the intuition behind the use of a selective attention mechanism for clustering. Now, we describe how this intuition is translated into a workable algorithm. We first focus on the determining a robust selection threshold (\ref{subsec2}), use this threshold to define the core structures (\ref{sec:obtain_core}), and assign them cluster labels (\ref{sec:assigning}). Then, we provide details on how full clusters are reconstructed from the retained cores (\ref{sec:reintegrating}). Finally, the algorithm pseudocode is presented (\ref{sec:algo}) along with an analysis of its time and space complexity (\ref{sec:complex}). 

\subsection{Determination of a pruning threshold}\label{subsec2}
To operationalize the concept of selective attention, we require a threshold-based mechanism that prunes irrelevant data points (i.e., focuses only on relevant data) during cluster formation. To identify isolated points $p^i=\{p^i_1, p^i_2, ..., p^i_n\}$ in an n-dimensional Cartesian space, pairwise distances should be computed. The original distance-matrix approach requires $O(N^2)$ time and memory which is not cost effective, so, to improve the computational efficiency of the proposed clustering algorithm, we replace the explicit construction of the full pairwise distance matrix with a spatial indexing structure based on the k-dimensional tree (KD-tree) \cite{bentley1975multidimensional}. The KD-tree enables efficient nearest-neighbor queries while avoiding the quadratic time and space complexity associated with storing all pairwise distances. 

A KD-tree is a binary tree that recursively partitions an n-dimensional space along the coordinate axes. Given a dataset $\mathcal{P} = \{p^1, p^2, \ldots, p^N\} \subset \mathbb{R}^n$, the tree is constructed by selecting a splitting dimension (typically alternating across dimensions or chosen based on variance) and dividing the dataset at the median value along that dimension. Once the KD-tree is built by Euclidean distance, nearest-neighbor queries can be performed efficiently. For each point ${p}^i$, we query the KD-tree to identify its nearest neighbor ${p}^j$, $j \neq i$, under the Euclidean metric. This replaces the need to compute all pairwise distances $d_{ij}$. The nearest-neighbor distance is therefore defined as 
                                    \begin{equation}
                                    r_i = \min_{j \ne i} \left\| p^i - p^j \right\|_2
                                     \label{eq:Equation3}
                                    \end{equation}
\noindent and build a vector of minimum distances $R = [r_i] = \big[r_1, r_2, \ldots, r_N\big]$. 

To ensure robustness, we utilize the Modified Z-score criterion \cite{iglewicz1993outliers} to exclude outliers from $R$. This criterion computes a Z-scores using the population median ($\tilde{\mu}$) and the median absolute deviation (MAD; $\tilde{\sigma}=median(|r_i-\tilde{\mu}|)$) as robust estimators of the mean and the standard deviation, respectively. Accordingly, raw \( r_i \) values are converted into Modified Z-scores as follows

\begin{equation}
    \textit{m}_i =  \frac{0.6745 \cdot (r_i - \tilde{\mu})}{\tilde{\sigma}},
    \label{eq:Equation10}
\end{equation}

\noindent where the constant 0.6745 is used to scale the MAD so that it is comparable (i.e., on the same scale) as the standard deviation~\cite{iglewicz1993outliers}. After computing the Modified Z-scores, we obtain a set of minimal distances $r_i$ excluding outliers

\begin{equation}
    \hat{R} = [\hat{r_i} \ \mid m_i  > \tau_{rej}] 
    \label{eq:R_hat}
\end{equation}
Empirically, we found that using a lenient threshold ($\tau_{rej} = 10$) for outlier rejection yields the best trade-off between sensitivity and specificity for SACA. This threshold allows preserving edge points while excluding aberrant distances.

Using this robust set of distances, we extract two data-driven properties:

\begin{equation}
L \;=\; \max_i \hat{r_i} .
\end{equation}
\begin{equation}
\sigma_{\text{opt}} = Q_{0.01}(\hat{R})
\label{eq:Equation4}
\end{equation}

\noindent where $Q_{0.01}(\hat{R})$ stands for the function that returns an estimate of quantile 0.01 for the distribution of $\hat{R}$. Using these definitions, \(L\) is a robust estimate of the largest nearest-neighbor distance. Thus, $L^{-1}$ is a proxy for the dataset's minimal density. Intuitively, we want \(\sigma_{\text{opt}}\) to be the smallest inter-point spacing (i.e., the overall nearest-neighbor distance; $\sigma_{\text{opt}}^{-1}$ is a proxy for the dataset's maximal density). However, since extrema values from  samples are known to be unreliable, we use $Q_{0.01}$ instead of the minimum as a robust estimator of the minimal inter-point spacing. In practice, using $Q_{0.01}$ helps to avoid the possibility of $\sigma_{\text{opt}}^{-1} \rightarrow \infty$ when two points in a datasets randomly happens to be on top of one another. Because the distribution of $r_i$ is expected to be right-skewed and bounded by 0, its left tail is compressed and very small differences in $\sigma_{\text{opt}}$ can cause huge differences in $\sigma_{\text{opt}}^{-1}$ as $\sigma_{\text{opt}} \rightarrow 0$. Such points would not be detected as outlier because of distributional asymmetry, motivating this additional precaution.
    
$L$ and $\sigma_{opt}$ are the key parameters and form the foundation of the proposed method. To define a density boundaries, using these data-driven values, a global threshold $T$ is determined as follows

\begin{equation}
T \;=\; \left\lceil \frac{L}{2\sigma_{\text{opt}}}\right\rceil.
\label{eq:Equation6}
\end{equation}
\noindent where $\left\lceil \  \right\rceil$ represent the ceiling function. Without the ceiling function, a slightly underestimated neighborhood radius can fragment continuous clusters into several smaller ones. We tested the practical impact of this ceiling on the algorithm performance and confirmed the importance of using this ceiling in Appendix 1. A more formal justification for the definition of $T$ can be found in Appendix 2. An additional conceptual motivation for the use of the threshold $T$, conceptualize in relation to the inter-cluster margin is presented in Appendix 3.

\subsection{Obtaining the core structure of clusters}\label{sec:obtain_core}

Having defined a pruning threshold $T$, we now explain how this threshold is used to prune the dataset and obtain the core structure of the clusters (i.e., the denser part of the clusters). First, for a point $p^i$, we normalize the distance of its neighbors $D_i = [d_{i1}, d_{i2}, ..., d_{iN}]$ as  $\frac{D_i}{2\sigma_{\text{opt}}}$. The factor \(1/2\) in this normalization compensates the same factor in the definition to $T$ and is further explained in Appendix 2. Then, we select all the neighbors within a ball of radius $T$ 

\begin{equation}
n_i \;=\; \arg\!\left(\frac{D_i}{2\sigma_{\text{opt}}} < T \right)
\label{eq:Equation7}
\end{equation}

Defining the weight of sample \(p^i\) as the cardinality of its selected neighbor set,
\begin{equation}
w_i = |n_i|,
\label{eq:equation8}
\end{equation}
we prune low-density samples using
\begin{equation}
\left\{
    \begin{array}{ll}
        \text{if } w_i \leq C & \text{prune the sample}, \\
        \text{else} & \text{use the sample for the core structure of the cluster},
    \end{array}
\right.
\end{equation}
where \(C\) is referred to as the Attention Selectivity Coefficient. This name is appropriate given that $C$ modulates whether the algorithm "pay attention" to more isolated points (i.e., points with smaller weight $w_i$) or focus only on points that are part of denser structures. With the core structures thus extracted, the next step involves organizing these dense regions into coherent clusters.

\subsection{Assigning clusters to the cores}\label{sec:assigning}
Following the acquisition of the core structures of the clusters, we proceed to assign each point to a cluster. This process begins by randomly selecting a sample from retained candidates and assigning it to the first cluster. Next, all its unpruned neighbors are examined and those that are not yet assigned to any cluster are added to the same cluster. These newly assigned points are then placed on a list for further examination. Each point on this list undergoes the same procedure and is subsequently removed once processed. When the list becomes empty, it indicates that the core structure of one cluster has been fully established. Another random candidate from the unclustered points is then selected, and the same process is repeated. Due to the the sufficient margin derived from previous phase (i.e. pruning phase) between these structures, it guarantees that cores do not overlap or connect with one another. Once all structures have been assigned to clusters, the final step is to reintegrate the pruned items into their appropriate clusters. 

\subsection{Reintegrating sparse points}\label{sec:reintegrating}

Once core structures have been associated to clusters, two methodologies are employed to gradually recreate the final clusters using the trimmed data samples. The first method assigns each data point to the cluster of its closest clustered neighbor. In the algorithm, this approach corresponds to setting the boolean parameter \texttt{use\_center} to \texttt{False} which is also the default configuration. The second method assigns data points based on the closest cluster core centroid. This centroid-based approach is particularly effective for convex datasets. It can be enabled by setting \texttt{use\_center} to \texttt{True}. The nearest-neighbor assignment (default) is recommended for arbitrary or non-convex shapes and noisy boundaries; the centroid-based assignment is recommended for convex clusters with significant overlap. In both cases, these assignment transforms the set of pruned cluster cores into the final clustering solution.

\subsection{Algorithm}\label{sec:algo}

SACA is designed to overcome common limitations of traditional density-based clustering methods, which often depend heavily on carefully tuned parameters and struggle with clusters of varying densities. By minimizing reliance on user-defined parameters, SACA operates in a more autonomous and user-friendly manner. As will be shown in our results, SACA performs well in most cases in its default configuration, requiring no parameter tuning. When initial clustering results are unsatisfactory, the Attention Selectivity Coefficient can be adjusted. Increasing $C$ limits the inclusion of sparsely distributed points when forming initial cluster cores, thereby focusing the algorithm on the fundamental structure of each cluster. These core structures correspond to the high-density regions of arbitrarily shaped clusters.

\begin{figure}[H]
    \centering
    \includegraphics[width=\textwidth]{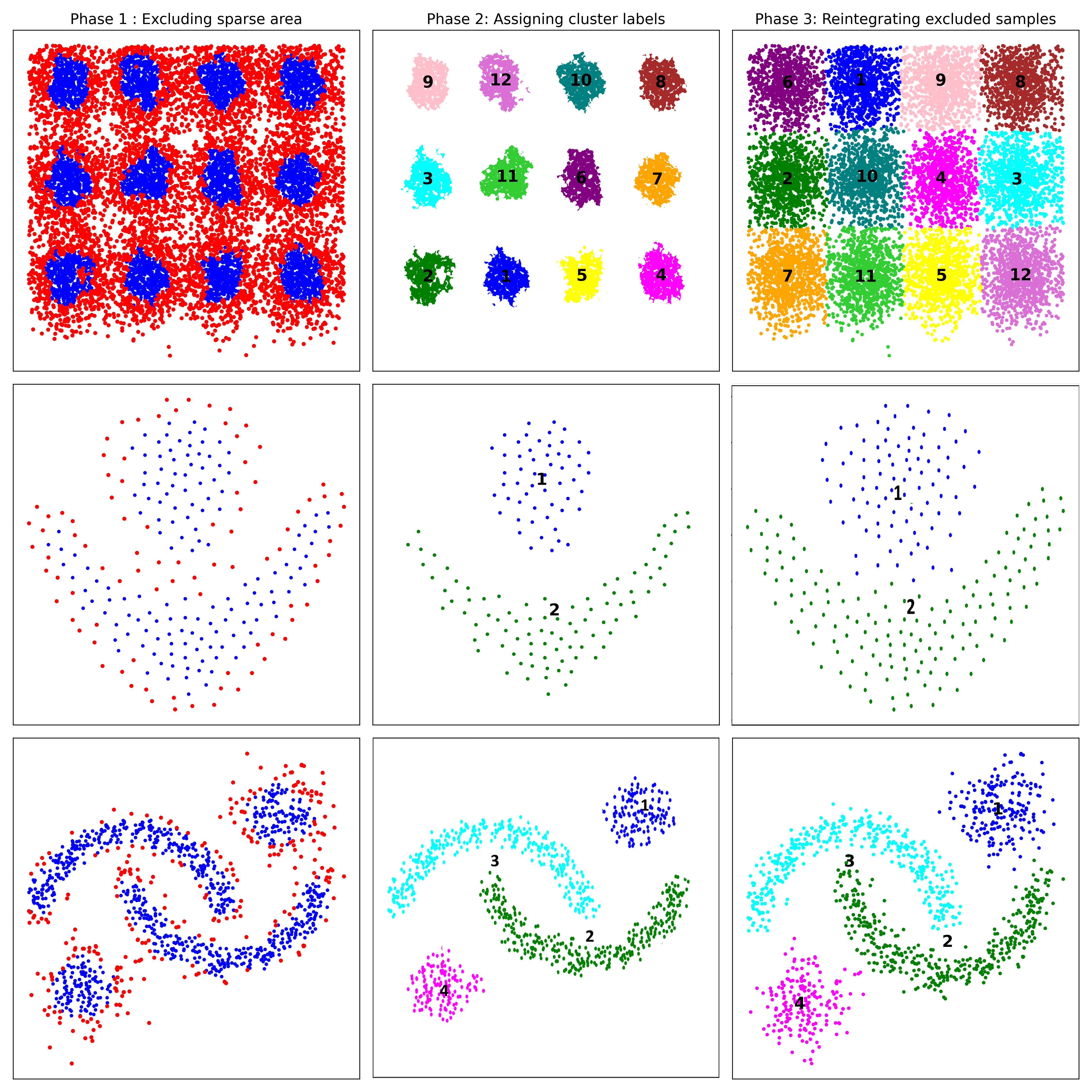}
    \caption{Graphical illustration of the method’s key steps (columns) using three examples of cluster distributions (rows).}
    \label{fig:Figure 2}
\end{figure}

To summarize the idea behind the algorithm, we can decompose its operation in three main phases (Fig.~\ref{fig:Figure 2}). In the first phase, SACA computes the full pairwise distance matrix and finds each point’s nearest-neighbor distance. It then filters out outlier distances (using a modified Z-score criterion) and computes a global threshold $T$ to distinguish high-density core points from potential noise (Section \ref{subsec2}). Using this threshold and the selectivity coefficient $C$, SACA identifies high-density regions and forms initial core clusters (Section \ref{sec:obtain_core}). In the second phase, it assigns cluster labels to the shapes emerged from the previous phase (Section \ref{sec:assigning}). Finally, in the third phase, the algorithm expands these clusters by reassigning the remaining (previously filtered in the first phase) points to their nearest core cluster (Section \ref{sec:reintegrating}). This assignment can be performed using cluster centroids or by nearest-neighbor linking, depending on whether the optional \texttt{use\_center} parameter is set to \texttt{True}. The result of this phase is the final outcome of the clustering. The pseudo-code in Algorithm \ref{algo:saca} provides an overview of the algorithm's logic and key operations.

\begin{algorithm}[H]
  \caption{SACA Clustering}
  \label{algo:saca}
  \begin{algorithmic}[1]
    \Procedure{SACA}{$C=1,\;use\_center=false$}

      \State \textbf{1. Initialize:}
      \State$N\gets|data|$; 
      $cluster\_num, added\_neighbor\gets 0^N$; 
      \State$noise,non\_noise,biglist\gets[\,]$
      \Statex

      \State \textbf{2. Building KD-tree, Find Outliers, Extract Threshold:}
     \State $tree\gets$ creating KD-tree; 
    \State $R \gets$  set of the nearest‐neighbor distances for all points from $tree$

    \State $outlier\_idx\gets\text{modified\_z\_score\_filter}(R)$ 
    \State $\hat{R} \gets $ remove entries at $outlier\_idx$ from $R$
    \State $\sigma_{opt}\gets Q_{0.01}({\hat{R}})$
    \State $L\gets\max(\hat{R})$
    \State $T\gets\lceil L/(2\sigma_{opt})\rceil$
    \Statex

    \State \textbf{3. Forming the core structure of clusters (Phase 1):}

    \State $neighborlist \gets \text{query\_ball\_point}(tree,\; data,\; r=T)$
    \For{$i = 0 \to N-1$}
    \If{$|neighborlist[i]| \le C$}
        \State $noise \gets noise \cup \{i\}$
    \Else
        \State $non\_noise \gets non\_noise \cup \{i\}$
    \EndIf
    \EndFor
     \For{each $i \in noise$} 
    \State $cluster\_number[i] \gets -1$
\EndFor
\Statex

      \State \textbf{4.  Assigning clusters to the cores (Phase 2):}
      \State $unclustered\gets|non\_noise|,\;current\_cluster\gets0$
      \While{$unclustered>0$}
        \State pick random $i$ with $cluster\_num[i]=0$
        \State $current\_cluster\gets current\_cluster+1$
        \State $neighbors\_queue\gets[i]$
        \While{$neighbors\_queue\neq\emptyset$}
          \State $p\gets neighbors\_queue.pop()$; $cluster\_num[p]\gets current\_cluster$
          \State $unclustered\gets unclustered-1$
          \ForAll{$q\in neighborlist[p]$}
            \If{$cluster\_num[q]=0$}
              \State $cluster\_num[q]\gets current\_cluster$
              \If{$added\_neighbor[q]=0$} push $q$; $added\_neighbor[q]\gets1$ \EndIf
            \EndIf
          \EndFor
        \EndWhile
      \EndWhile
      \Statex
  
      \State \textbf{5. Reintegrating the points in sparse areas to clusters (Phase 3):}
      \If{$use\_center$}
        \State compute cluster centers; assign each noise point to nearest center
      \Else
        \State assign each noise point to its nearest non-noise neighbor’s cluster
      \EndIf

    \EndProcedure
  \end{algorithmic}
\end{algorithm}

\FloatBarrier

\subsection{Time and Space Complexity Analysis}\label{sec:complex}

Given data $X \in \mathbb{R}^{N \times d}$, where $d$ is the dimensionality of each sample, $N$ is the number of samples, and $C$ is the number of clusters found (with $C \leq N$), the space and time complexity of the SACA algorithm are described as below.

\vspace{5pt}

\noindent \textbf{Space Complexity:} The KD-tree requires $O(N)$ memory to store the dataset points along with the tree structure, where each node typically contains a data point and references to child nodes. Unlike the full distance matrix approach, which requires $O(N^2)$ memory to store all pairwise distances, the KD-tree has linear memory requirements. Let $E$ denote the total number of neighbors across all samples within the T-neighborhood, where $E \ll N^2$. Storing these neighborhood relationships requires an additional $O(E)$ memory. Therefore, the overall space complexity of the algorithm is $O(N + E)$.

\vspace{5pt}
\noindent \textbf{Time Complexity:} The construction of a KD-tree from $N$ points in $\mathbb{R^n}$ typically requires $(N\log N)$ time when median splitting is used at each level to ensure a balanced tree. Once constructed, nearest-neighbor queries can be performed in average $O(logN)$ time per query due to the tree’s hierarchical partitioning and effective pruning via branch-and-bound search. Therefore, computing all nearest-neighbor distances ${r_i}$ for the dataset requires $O(N\log N)$ time.

\section{Experiments}

\subsection{Datasets}

To test our algorithm, we reused different artificially created benchmark datasets with varying shapes and densities \cite{Franti2018, Barton2023}. We also created additional datasets to assess specific situations (Table \ref{tab:datasets}).
                                                                
\begin{table}[ht]
\centering
\footnotesize
\caption{\normalsize Datasets used for benchmarking.}
\label{tab:datasets}
\begin{tabular}{llccc}
\toprule
\textbf{\small Dataset} & \textbf{ \small Source} & \multicolumn{3}{c}{\textbf{\small Attributes}} \\
\cmidrule(lr){3-5}
 &  & Samples & Features & Clusters \\
\midrule
Spiral                & \cite{Franti2018} & 312   & 2 & 3  \\
Birch1 (12 clusters)  & \cite{Franti2018} & 11872 & 2 & 12 \\
A3                    & \cite{Franti2018} & 7500  & 2 & 50 \\
S2                    & \cite{Franti2018} & 5000  & 2 & 15 \\
Flame                 & \cite{Franti2018} & 240   & 2 & 2  \\
Skewed                & \cite{Franti2018} & 1000  & 2 & 6  \\
Cluto-t4-8k           & \cite{Barton2023} & 8010  & 2 & 6  \\
Complex-9             & \cite{Barton2023} & 3040  & 2 & 9  \\
Smile 2               & \cite{Barton2023} & 1010  & 2 & 4  \\
DS-850                & \cite{Barton2023} & 850   & 2 & 5  \\
Noisy Circles         & Author made       & 2400  & 2 & 7  \\
Rings                 & Author made       & 7200  & 2 & 36 \\
Noisy Spiral          & Author made       & 800   & 2 & 2  \\
Moons-Stars           & Author made       & 1200  & 3 & 4  \\
3Compound             & Author made       & 1500  & 2 & 3  \\
Unbalanced            & Author made       & 1500  & 2 & 6  \\
\bottomrule
\end{tabular}
\end{table}

\FloatBarrier

\subsection{Evaluation Metrics}

To accurately measure the performance of the clustering algorithm, we calculated six established clustering evaluation metrics.

The \textbf{Silhouette index} assesses clustering performance by measuring the difference between distances within clusters and between clusters. For a sample $i$, let $a(i)$ be the average intra-cluster distance and $b(i)$ be the minimum average distance to points in other clusters. The Silhouette score is defined as
\begin{equation}
s(i) = \frac{b(i) - a(i)}{\max\{a(i), b(i)\}}.
\end{equation}
Values range from $-1$ to $+1$, with higher values indicating better clustering quality \cite{rousseeuw1987silhouettes}.

The \textbf{Calinski-Harabasz (CH) index} evaluates cluster validity by comparing between-cluster and within-cluster dispersion:
\begin{equation}
\text{CH} = \frac{\mathrm{Tr}(B_k)/(k-1)}{\mathrm{Tr}(W_k)/(N-k)},
\end{equation}
where $\mathrm{Tr}(B_k)$ and $\mathrm{Tr}(W_k)$ denote between-cluster and within-cluster scatter, respectively, $k$ is the number of clusters, and $N$ is the number of samples. Higher values indicate better clustering \cite{wang2019improved}.

The \textbf{Davies-Bouldin (DB) index} measures the ratio of within-cluster scatter to between-cluster separation:
\begin{equation}
\text{DB} = \frac{1}{k} \sum_{i=1}^{k} \max_{j \neq i} \left( \frac{S_i + S_j}{M_{ij}} \right),
\end{equation}
where $S_i$ is the average distance of points in cluster $i$ to its centroid, and $M_{ij}$ is the distance between centroids of clusters $i$ and $j$. Lower values indicate better clustering \cite{article11}.

When ground truth labels are available, additional external validation metrics can be used. The \textbf{Adjusted Rand Index (ARI)} measures similarity between two clustering results based on pair counting. Let $a$ denote the number of pairs assigned to the same cluster in both true and predicted labels, $b$ the number of pairs assigned to the same cluster in the true labels but to different clusters in the predicted labels, $c$ the number of pairs assigned to different clusters in the true labels but to the same cluster in the predicted labels, and $d$ the number of pairs assigned to different clusters in both cases. The ARI is defined as
\begin{equation}
\text{ARI} = \frac{2(ad - bc)}{(a + b)(b + d) + (a + c)(c + d)}.
\end{equation}
The ARI equals 0 for random clustering and 1 for perfect agreement \cite{hubert1985comparing}.

The \textbf{Adjusted Mutual Information (AMI)} measures agreement between two clusterings while correcting for chance:
\begin{equation}
\text{AMI} = \frac{I(U,V) - \mathbb{E}[I(U,V)]}{\max\{H(U), H(V)\} - \mathbb{E}[I(U,V)]},
\end{equation}
where $I(U,V)$ is the mutual information between clusterings $U$ and $V$, $H(\cdot)$ denotes entropy, and $\mathbb{E}[I(U,V)]$ is the expected mutual information. Values range from 0 (random agreement) to 1 (perfect match) \cite{vinh2009information}.

The \textbf{Completeness metric} evaluates whether all data points belonging to the same class are assigned to a single cluster:
\begin{equation}
\text{Completeness} = 1 - \frac{H(U \mid V)}{H(U)},
\end{equation}
where $H(U \mid V)$ is the conditional entropy of the true labels $U$ given predicted clusters $V$, and $H(U)$ is the entropy of the true labels. Values range from 0 to 1, with 1 indicating perfectly complete clustering \cite{rosenberg2007v}.

\section{Results}

\subsection{Overall Performances}

\begin{figure}[!htbp] 
    \centering
    \includegraphics[width=1.0\textwidth]{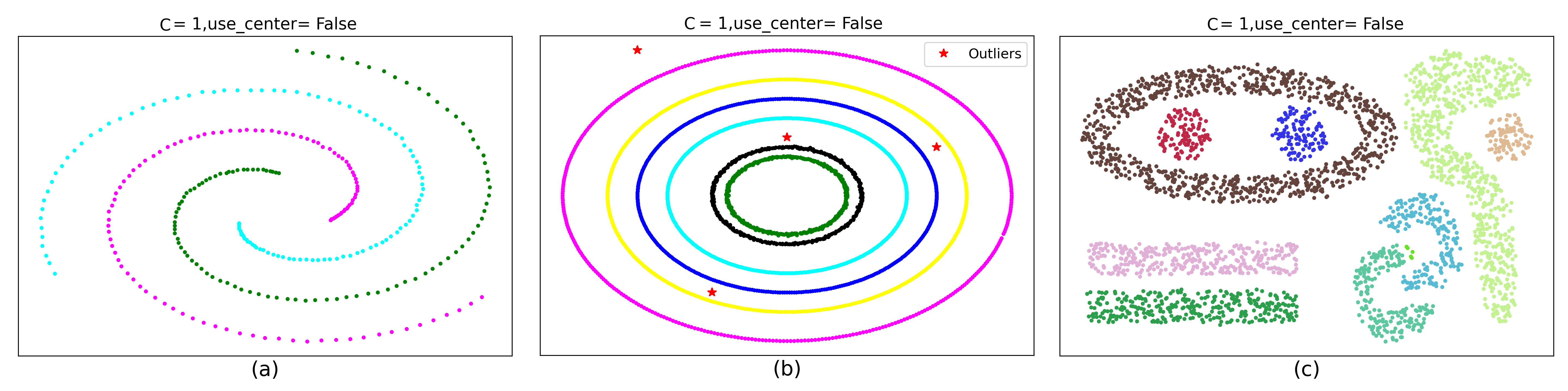}     \caption{Clustering results in the context of uniform cluster densities (using default parametrization): (a) Spiral, (b) Noisy Circles, (c) Complex-9.}
    \label{fig:Figure 3}
\end{figure}

\begin{figure}[!htbp]
    \centering
        \includegraphics[width=0.9\textwidth]{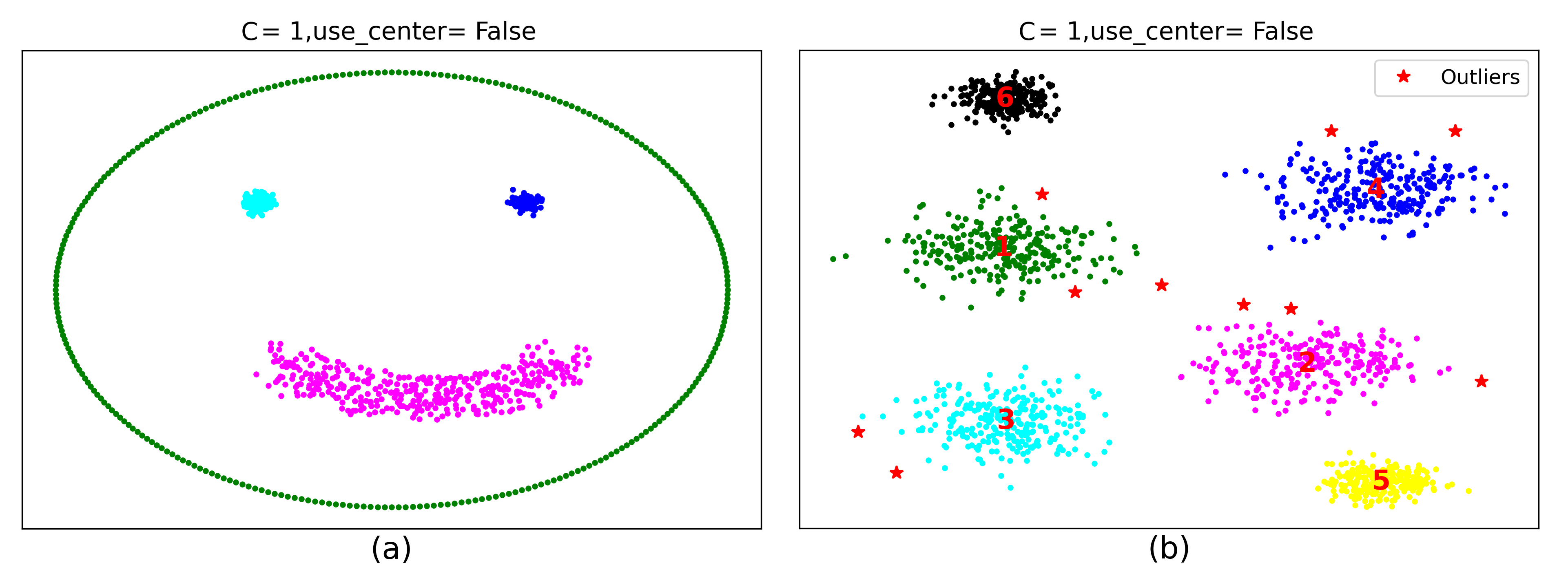}     \caption{Clustering results in presence of non-uniform cluster densities (using default parametrization): (a) Smile 2 and (b) Unbalanced.}
    \label{fig:Figure 4}
\end{figure}

\begin{figure}[!htbp]
    \centering
   \includegraphics[width=0.8\textwidth]{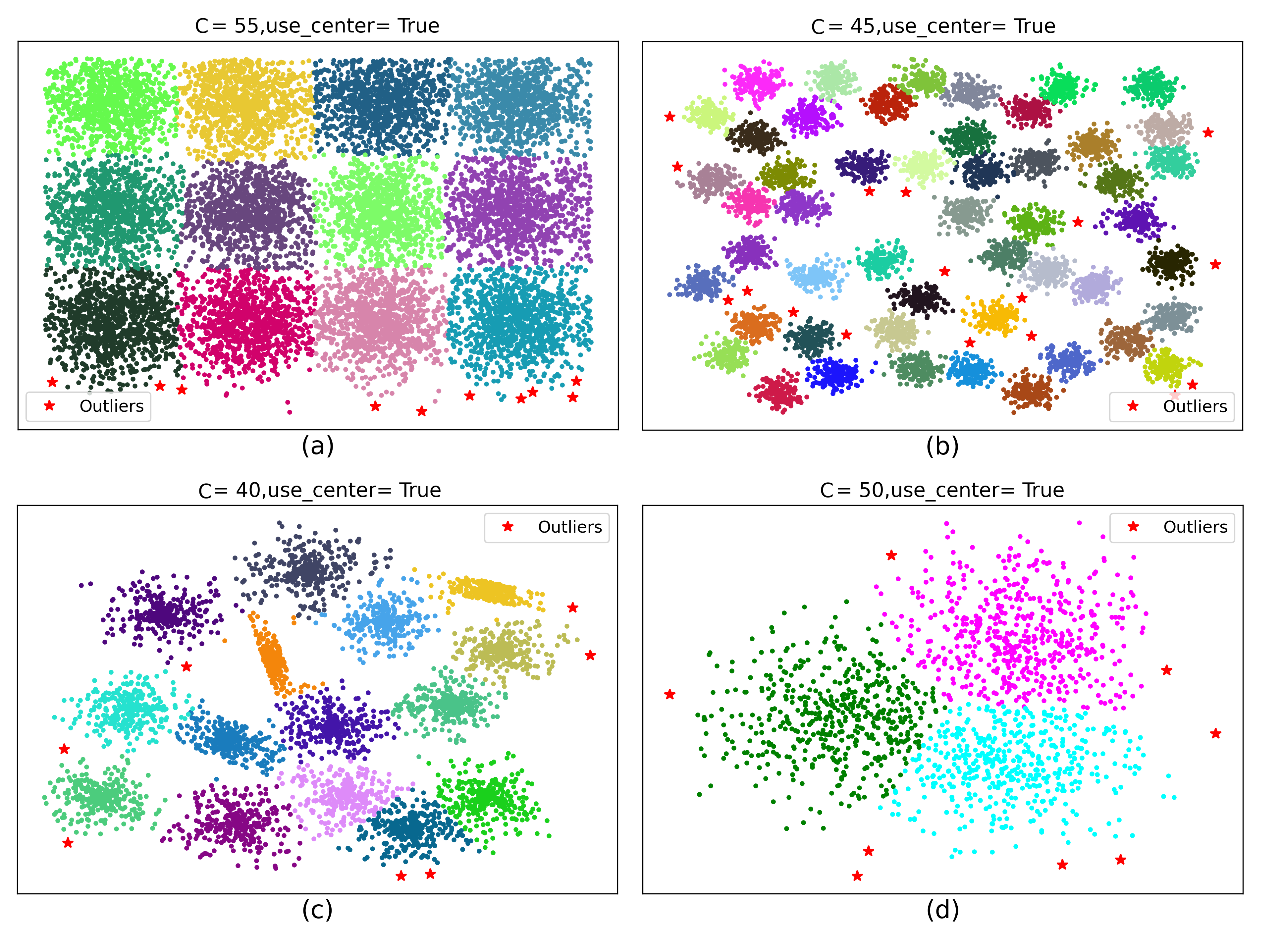}     \caption{ Clustering results on clusters with conflicts: (a) Birch1, (b) A3, (c) S2, and (d) 3-Compound.}
      \label{fig:Figure 5}
\end{figure}

\begin{figure}[!htb]
    \centering
    \includegraphics[width=0.8\textwidth]{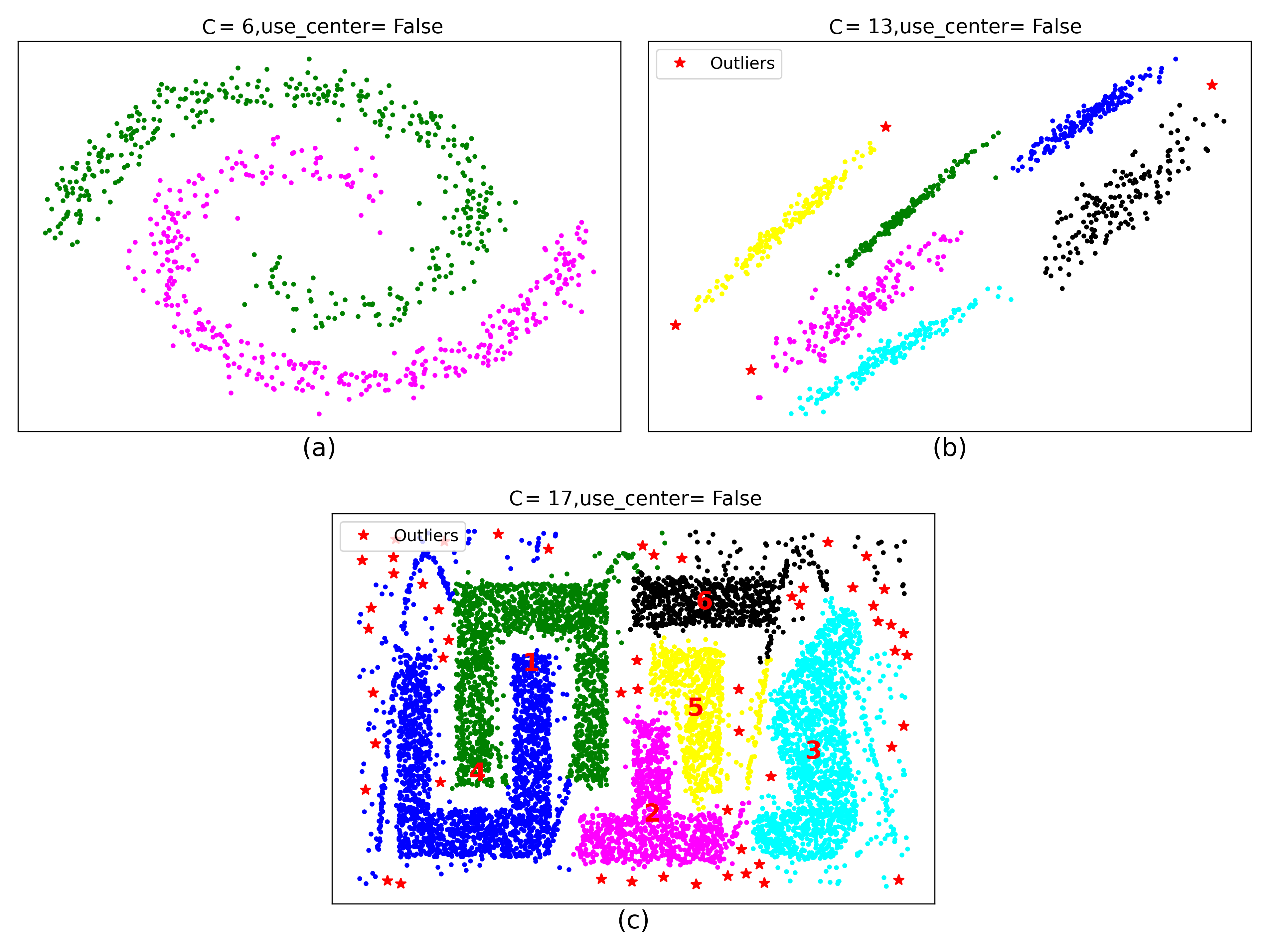}
    \caption{Clustering results on challenging, noisy clusters without conflicts : (a) Noisy Spiral, (b) Skewed, (c) Cluto-t4-8k.}
       \label{fig:Figure 6}
\end{figure}

\begin{figure}[!htb]
    \centering
    \includegraphics[width=0.9\textwidth]{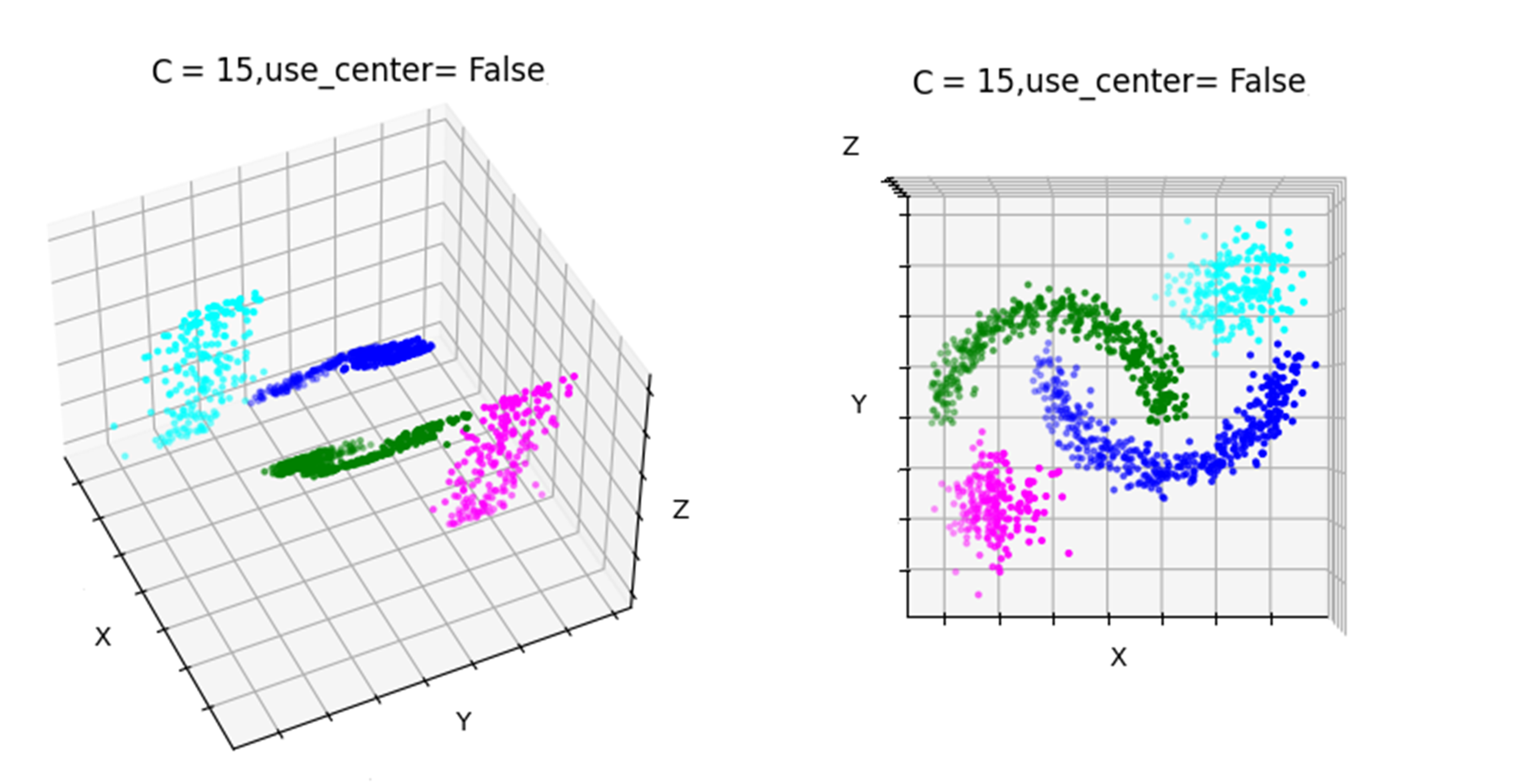}
    \caption{Clustering result in 3D using the Moons-Stars dataset.}
       \label{fig:Figure 7}
\end{figure}

In this section, we evaluate the performance of the algorithm when applied to different dataset configurations. Table~\ref{tab:saca_fig49}  lists the parameter settings used for applying the method to the datasets shown in  Figs.~\ref{fig:Figure 3} to \ref{fig:Figure 8} and described in Table~\ref{tab:datasets}. The values C=1 and \verb|use_center = False| are used as default values, with no user involvement in setting the parameter.

\begin{table}[h]
\centering
\footnotesize 
\setlength{\tabcolsep}{3.5pt} 
\caption{\small Parameter settings of SACA used in Figures 4--9. Non default values are shown in bold.}
\label{tab:saca_fig49}
\begin{tabular}{llcc}
\toprule
\footnotesize
\textbf{Figure} & \textbf{Dataset(s)} & \textbf{C} & \textbf{use\_center} \\
\midrule
\multirow{3}{*}{Fig.~4}  
 & Spiral & 1 & False \\
 & Noisy Circles & 1 & False \\
 & Complex-9 & 1 & False \\
\midrule
\multirow{2}{*}{Fig.~5} 
 & Smile 2 & 1 & False \\
 & Unbalanced & 1 & False \\
\midrule
\multirow{4}{*}{Fig.~6}  
 & Birch1 & \textbf{55} & \textbf{True} \\
 & A3 & \textbf{45} & \textbf{True} \\
 & S2 & \textbf{40} & \textbf{True} \\
 & 3-Compound & \textbf{50} & \textbf{True} \\
\midrule
\multirow{3}{*}{Fig.~7}  
 & Noisy Spiral & \textbf{6} & False \\
 & Skewed & \textbf{13} & False \\
 & Cluto-t4-8k & \textbf{17} & False \\
\midrule
\multirow{1}{*}{Fig.~8}  
 & Moons-Stars (3D) & \textbf{15} & False \\
\midrule
\multirow{2}{*}{Fig.~9}    
 & Rings (global level) & 1 & False \\
 & Rings (local level) & \textbf{20} & False \\
\bottomrule
\end{tabular}
\begin{flushleft}
\tiny
\end{flushleft}
\end{table}
\FloatBarrier

\FloatBarrier

The algorithm performs optimally with default parameters when applied to datasets with uniform cluster densities (Fig. \ref{fig:Figure 3}). Furthermore, it yields reliable results for datasets with groups of varying densities or noise, provided that condition (\ref{eq:Margin_T}) (Appendix 3) is met. Fig.~\ref{fig:Figure 4} illustrates two datasets with different cluster densities. In panel (b), for instance, six clusters are present, with clusters 5 and 6 having densities twice as high as clusters 1 to 4. The accuracy of the algorithm in this case is due to the satisfaction of condition (\ref{eq:Margin_T}) despite the density differences.

\FloatBarrier

Another scenario occurs when the condition (\ref{eq:Margin_T}) is not met and there is overlap between clusters. In this case, the user should adjust the value of C. Increasing its value enhances pruning effectiveness and results in smaller, more detailed clusters, while decreasing it produces fewer, broader clusters. This parameter effectively functions as the "level of attention" for the algorithm and enables precise control over the clustering to meet specific analysis requirements. See Figs. \ref{fig:Figure 5} to \ref{fig:Figure 7} for examples of the effect of varying $C$ on different cluster configurations. The datasets used in \ref{fig:Figure 5} consist of convex-shaped clusters with high conflict; therefore, the parameter \texttt{use\_center} was set to True.

\FloatBarrier

\FloatBarrier

\FloatBarrier

\subsection{Multilevel Pattern Recognition}
Datasets often exhibit diverse structures, where each pattern or cluster may be composed of smaller sub-clusters with varying geometries. The ability to accurately identify these sub-clusters without the necessity of precise parameter tuning is a crucial advantage for any clustering algorithm. The proposed method offers this flexibility, allowing the identification of various patterns within the dataset through the simple adjustment of \textit{C}.

\begin{figure}[!htb]
    \centering
    \includegraphics[width=0.9\textwidth]{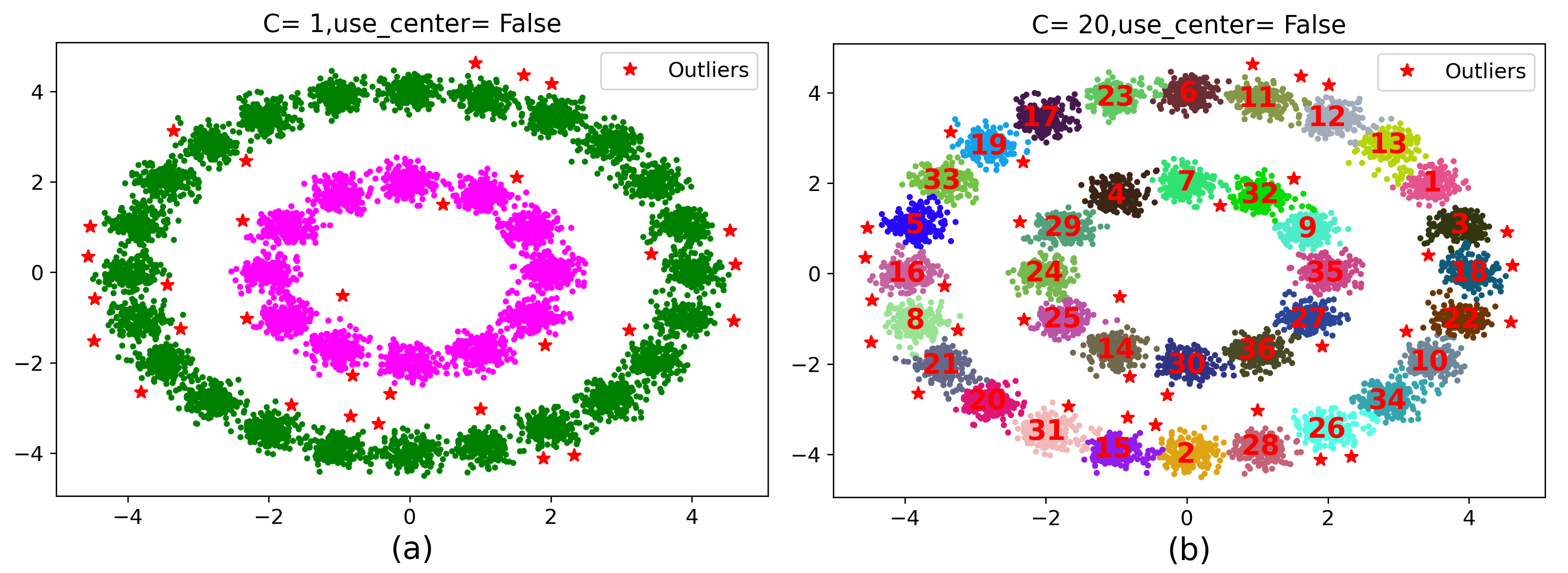}
    \caption{Rings made of small clusters: (a) result without user-defined C, (b) result with C=20. }
    \label{fig:Figure 8}
\end{figure}

\FloatBarrier

In panel (a) of Fig.~\ref{fig:Figure 8}, the algorithm, using the default $C=1$ setting, successfully identifies the two main clusters, represented by two large rings. However, these two main clusters are composed of smaller sub-clusters. By increasing $C$ to 20, the algorithm also detects all these smaller sub-clusters. This demonstrates that the proposed algorithm can easily uncover structures at multiple scales by adjusting $C$.

\subsection{Parameter analysis and settings}

\begin{figure}[!htb]
    \centering
    \includegraphics[width=1\linewidth]{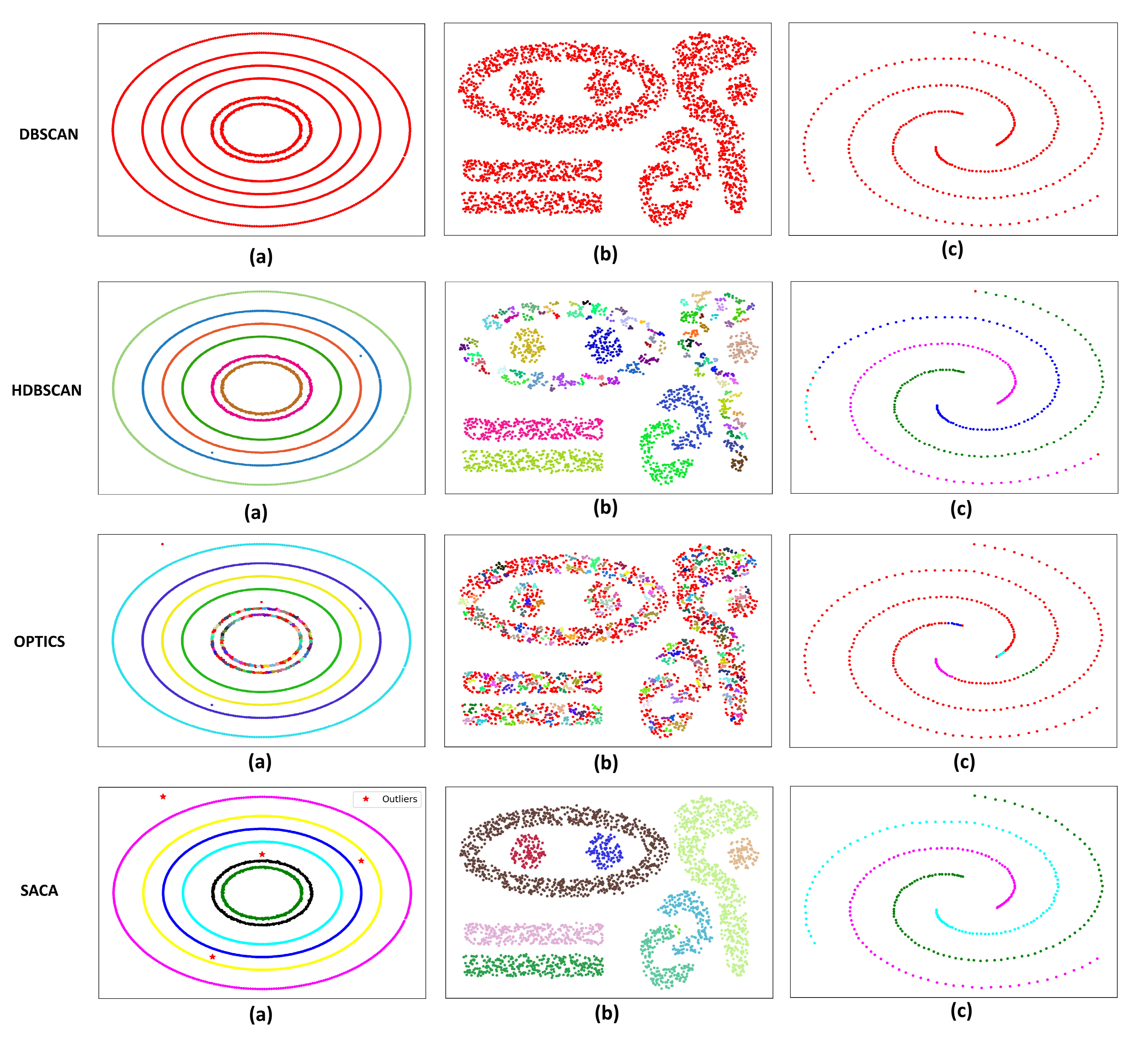}
    \caption{Performance with default parameters: (a) Circles, (b) complex 9, (c) Spiral }
    \label{fig:default-params}
\end{figure}

To evaluate the performance of SACA, we performed two experiments: (i) \textit{default, no tuning}; and (ii) \textit{best-tuned}. For the \textit{default, no tuning} experiment, all baseline algorithms (DBSCAN, OPTICS, and HDBSCAN) are evaluated using the unmodified scikit-learn configurations (i.e., without manual hyper-parameter tuning), and SACA is executed with its default settings. For this experiment, we chose three datasets that contains diverse shapes (Noisy Circles, Complex-9, and Spiral; Fig.~\ref{fig:default-params}). This experiment demonstrate that, for this kind of datasets, SACA can perform with default parametrization while these three alternative algorithms cannot. 

\begin{figure}[!ht] 
    \centering
    \includegraphics[width=0.7\linewidth]{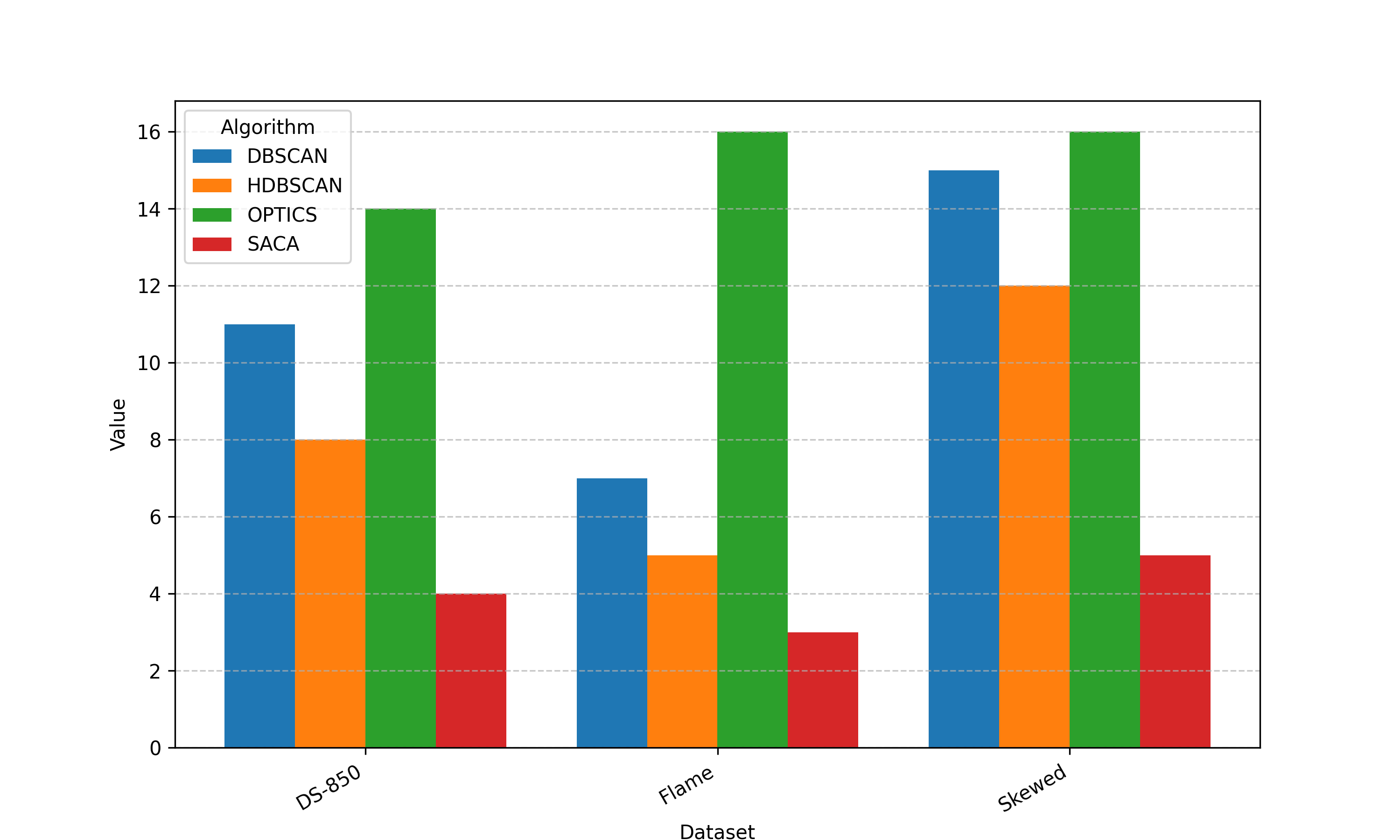}
    \caption{ Number of parameter adjustments to achieve results in Figure 11}
    \label{fig:adjustment-count}
\end{figure}

\begin{figure}[!ht] 
    \centering
    \includegraphics[width=0.8\linewidth]{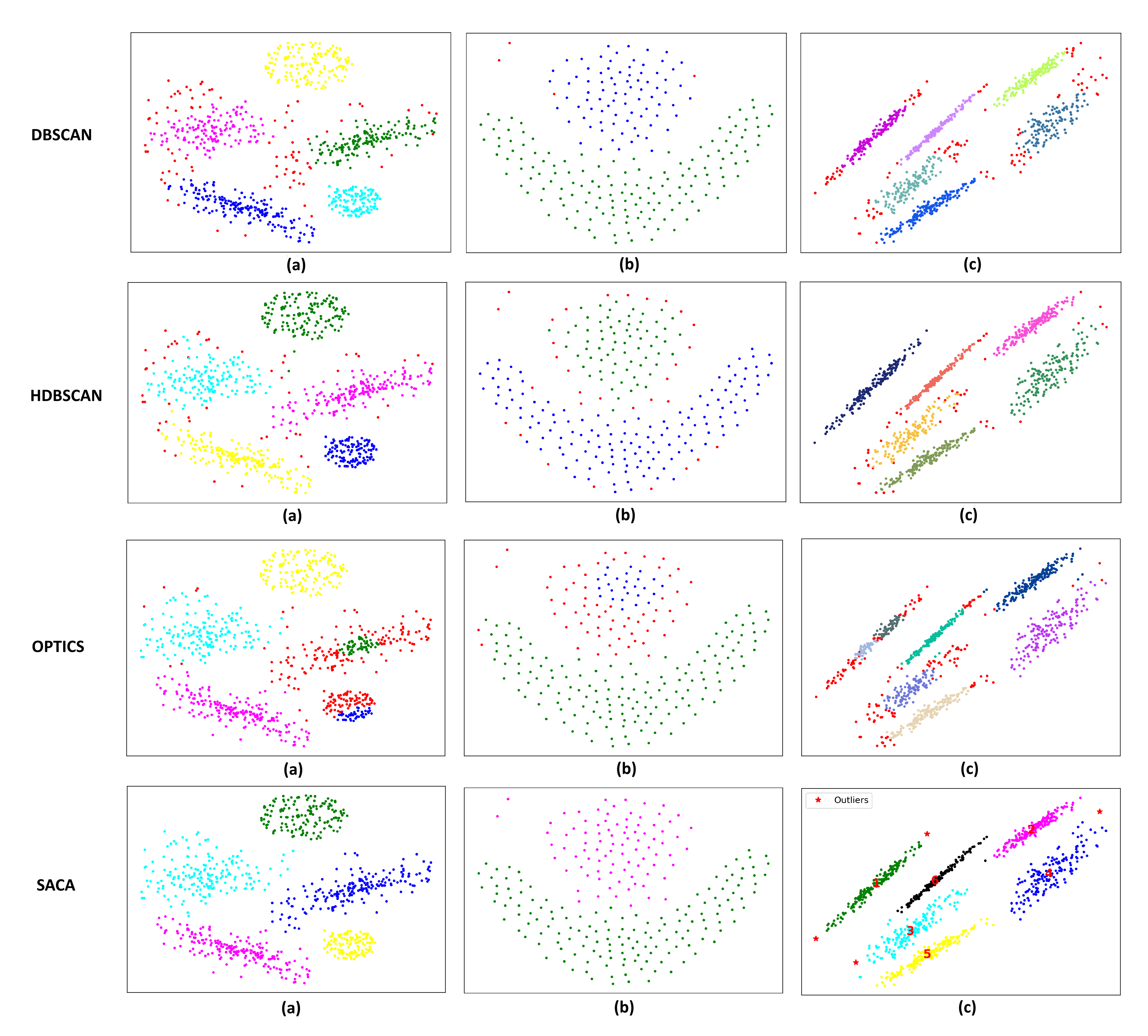}
    \caption{The outcomes of the best-tuned experiment: (a) DS-850, (b) Flame, (c) Skewed}
    \label{fig:best-tuned}
\end{figure}

In the \textit{best-tuned} experiment, each method is allowed to have targeted hyper-parameter changes to reach its strongest result. For this experiment, we chose three labeled benchmarks with distinct structure and density patterns (DS-850, Flame, and Skewed). These datasets best illustrate the need for fine tuning for both SACA and the baseline algorithms. The final (optimal) settings are listed in Table~\ref{tab:parameter_settings}, and the resulting scores for evaluation metrics are shown in Table~\ref{tab:performance_metrics}. To assess how much effort is required from the practitioner to use each of these algorithms, we report the number of adjustments necessary to optimize their outcome through manual fine-tuning. Each discrete parameter change (including toggling \texttt{use\_center}) is counted as one step, starting from the library default for baselines (Fig.~\ref{fig:adjustment-count}). For completeness, the best-tuned visual outcomes are shown in Fig.~\ref{fig:best-tuned}.

\FloatBarrier

For comparison without adjustments (Fig.~\ref{fig:default-params}), SACA performs competitively or better than DBSCAN, HDBSCAN, and OPTICS in the three shape-diverse datasets without manual hyperparameter tuning.

\begin{table}[h]
    \centering
    \footnotesize
    \caption{Parameter settings for best-tuned results.}
    \label{tab:parameter_settings}
    \begin{tabular}{llccc}
        \toprule
        \textbf{Algorithm} & \textbf{Parameter} & \multicolumn{3}{c}{\textbf{Datasets}} \\
        \cmidrule(lr){3-5}
         &  & DS-850 & Flame & Skewed \\
        \midrule
        \multirow{2}{*}{DBSCAN} 
         & $\varepsilon$ & 0.05 & 0.10 & 0.04 \\
         & $MinPts$ & 10 & 10 & 15 \\
        \midrule
        \multirow{2}{*}{HDBSCAN} 
         & $Mpts$ & 10 & 13 & 25 \\
         & $min\_cluster\_size$ & 10 & 4 & 9 \\
        \midrule
        \multirow{2}{*}{OPTICS} 
         & $Min\_sample$ & 25 & 10 & 30 \\
         & $\varepsilon$ & 1 & 5 & 10 \\
        \midrule
        \multirow{2}{*}{SACA} 
         & $C$ & 8 & 13 & 13 \\
         & $Use\_center$ & False & False & False \\
        \bottomrule
    \end{tabular}
\end{table}

\FloatBarrier

The number of parameter adjustments required to reach the best-tuned result for each algorithm dataset pair is shown in Fig.~\ref{fig:adjustment-count}. SACA reaches its best performance with fewer adjustments than baselines.

Fig.~\ref{fig:best-tuned} shows the best-tuned outcomes. SACA recovers the target cluster structures across the three benchmarks, while the baseline algorithms exhibit residual misassignments in challenging regions.

\FloatBarrier

As seen in Table~\ref{tab:performance_metrics}, SACA clearly outperforms the other three clustering algorithms in the evaluation metrics mentioned above. In particular, the results of the other three algorithms were obtained through iterative trial-and-error, whereas our algorithm achieved the best result in fewer attempts, as demonstrated by Fig. \ref{fig:adjustment-count}.

\begin{table}[!ht]
\centering
\footnotesize
\caption{\small The clustering performance metrics on three datasets.}
\label{tab:performance_metrics}
\begin{tabular}{llcccccc}
\toprule
\scriptsize
\textbf{Algorithm} & \textbf{Dataset} & \multicolumn{6}{c}{\textbf{Metrics}} \\
\cmidrule(lr){3-8}
 &  & Silh   $\uparrow$& CH   $\uparrow$& DB   $\downarrow$& ARI   $\uparrow$& AMI   $\uparrow$& Comp.   $\uparrow$\\
\midrule
\multirow{3}{*}{DBSCAN}  
 & DS-850 & 0.500 & 803.1 & 1.56 & 0.851 & 0.883 & 0.840 \\
 & Flame  & 0.257 & 59.5  & 1.55 & 0.940 & 0.880 & 0.840 \\
 & Skewed & 0.360 & 593.6 & 2.18 & 0.850 & 0.870 & 0.830 \\
\midrule
\multirow{3}{*}{HDBSCAN} 
 & DS-850 & 0.502 & 726.0 & 1.70 & 0.857 & 0.879 & 0.838 \\
 & Flame  & 0.197 & 41.9  & 2.95 & 0.760 & 0.680 & 0.580 \\
 & Skewed & 0.382 & 916.5 & 1.67 & 0.930 & 0.930 & 0.900 \\
\midrule
\multirow{3}{*}{OPTICS}  
 & DS-850 & 0.315 & 727.1 & 1.14 & 0.767 & 0.834 & 0.820 \\
 & Flame  & 0.158 & 64.2  & 1.97 & 0.790 & 0.710 & 0.620 \\
 & Skewed & 0.338 & 684.1 & 1.59 & 0.750 & 0.810 & 0.770 \\
\midrule
\multirow{3}{*}{SACA}    
 & DS-850 & \textbf{0.570}& \textbf{1517.4}& \textbf{0.53}& \textbf{0.986}& \textbf{0.983}& \textbf{0.983}\\
 & Flame  & \textbf{0.328}& \textbf{110.4}& \textbf{1.16}& \textbf{1.0}& \textbf{1.0}& \textbf{1.0}\\
 & Skewed & \textbf{0.395}& \textbf{1528.4}& \textbf{1.06}& \textbf{0.990}& \textbf{0.990}& \textbf{0.990}\\
\bottomrule
\end{tabular}
\end{table}

\FloatBarrier

\section{Discussion}

Across diverse shapes, both in the default (Fig.~\ref{fig:default-params}) and best-tune settings  (Tables~\ref{tab:parameter_settings}–\ref{tab:performance_metrics}, Figs.~\ref{fig:adjustment-count}–\ref{fig:best-tuned}), SACA is competitive with, or stronger than DBSCAN, HDBSCAN, and OPTICS while requiring fewer parameter changes. SACA works best when clusters exhibit clear inter-cluster margins even with differing densities and when non-convex geometries (rings, chains, spirals) are present (Figs.~\ref{fig:Figure 3},\ref{fig:Figure 4}, \ref{fig:Figure 6}). The global threshold \(T\) cuts off unwanted links and isolates high-density cores. Afterwards, reintegration attaches the pruned points to the correct cores. For approximately convex, partially overlapping roughly convex cluster structures, enabling \texttt{use\_center} improves edge assignments by shortening reintegration paths. From a practitioner’s point of view, the method is parameter-light with a simple rule. Start with the default settings; if you observe overlap, increase C in small increments (e.g., by 2–5). For denser datasets, use larger increments; for sparser datasets, use smaller ones. Keep \texttt{use\_center}{=}\texttt{False} for non-convex manifolds; set it to \texttt{True} for overlapping convex blobs. These rules reproduce the behaviors seen in Fig.~\ref{fig:Figure 3}- \ref{fig:Figure 8} . Sensitivity analyses indicate that the default outlier screening (Modified Z-score with a lenient cut-off) reduces spurious minima in the $\hat{R}$ histogram before computing \(T\), protecting the pruning step from distant anomalies while preserving edge points. Empirically, performance plateaus quickly as \(C\) increases, consistent with the low adjustment counts in Fig.~\ref{fig:adjustment-count}.

One main challenges remain: Uniform background noise over a wide range can inflate \(L\) and weaken the contrast that \(T\) provides; tightening the outlier screen on $R$ or raising \(C\) slightly reduces leakage from sparse regions. 

We also considered potential threats to validity. Implementation bias was controlled by using unmodified scikit-learn defaults for the baseline (“default”) runs and by recording the parameters selected for each dataset in the best-tuned runs (Table~\ref{tab:parameter_settings}). To address tuning-effort bias, we used the\emph{ adjustment count} (Fig.~\ref{fig:adjustment-count})  as a measure of human intervention. However, although we made our best effort to be fair in our fine-tuning attempts, since we could not blind the practitioner with respect to the identity of the algorithm being fine-tuned, we cannot exclude the possibility of inadvertent biases during the manual fine-tuning experiment. We acknowledge this limitation. Metric bias was mitigated by reporting six complementary performance metrics  (Table~\ref{tab:performance_metrics})  and cross-validating them with visual results (Fig.~\ref{fig:best-tuned}). 

Overall, for practitioners SACA offers a simple workflow. For researchers, the results suggest that combining a data-derived global threshold with late reintegration is a competitive alternative to the radius/min-samples approach used by algorithms such as DBSCAN, and OPTICS, particularly for non-convex structures. Future work will target adaptive schemes for uniform background noise fields while preserving SACA’s interpretability.

\section{Conclusion}

In this work, we introduced SACA, a novel density-based clustering method inspired by the concept of visual selective attention. By automatically determining a global neighborhood threshold from pairwise distances, SACA removes the burden of extensive parameter tuning that is common in traditional clustering algorithms. For more complex datasets, a single parameter, $C$, offers an intuitive way to adjust the level of clustering detail. SACA also introduces an optional toggle \textit{use\textbf{\_}center} parameter, which enables two flexible point-assignment strategies for the reintegration step: nearest-neighbor (default) and centroid-based, the latter being particularly effective for convex datasets. Our experiments demonstrate that SACA consistently outperforms well-known algorithms such as DBSCAN, HDBSCAN, and OPTICS in both accuracy and robustness. Beyond its strong performance, SACA provides practical benefits, including effective outlier handling, multilevel pattern discovery, and flexible assignment strategies through \textit{use\_center}. Although its performance on data dominated by uniformly distributed noise remains a challenge, we plan to address this through adaptive strategies.

\subsection*{Acknowledgments}

\section*{Declarations }

\begin{itemize}

\item Conflict of interest/Competing interests   All authors declare that they have no known competing financial interests or personal relationships that could have appeared to influence the work reported in this paper. 
\item Ethics approval and consent to participate  Not applicable
\item Consent for publication  Not applicable
\item Data availability   Benchmark datasets used in this study are publicly available from the original sources cited in Table 2 of the manuscript. The synthetic datasets generated during this study are available from the authors upon request and will be deposited in a public repository after acceptance. 
\item Materials availability   Not applicable
\item Code availability  The code supporting this study is not shared during peer review to protect intellectual property. It will be released in a public repository after acceptance. 
\item Author contribution    Meysam Shirdel Bilehsavar: conceptualization, methodology, implementation, data analysis, writing. Christian O'Reilly: conceptualization, data analysis, writing. Razieh Ghaedi: data collection, writing. 
\end{itemize}

\noindent

\bibliography{sn-bibliography}

\appendix 

\section*{Appendix 1} 

\textbf{Effect of Removing the ceiling function in the Calculation of $T$  } 

\noindent In this Appendix, we ran an experiment to justify empirically (see Appendix 2 for a more theoretical justification) the use of the ceiling functions in the definition of $T$. In this test, we removed the ceiling operation, modifying the original computation of $T$ as defined in (\ref{eq:Equation6}). Further, without the ceiling function, the $\frac{1}{2\sigma_{opt}}$ factor on both side of the inequality (\ref{eq:Equation7}) simplifies so that we have a modified threshold:
\[
T = \text{L}
\]
and a modified neighbor selection rule

\[
n_i \;=\; \arg\!\left(D_i < T \right)
\]

\begin{figure}[h]
    \centering
    \includegraphics[width=1\textwidth]{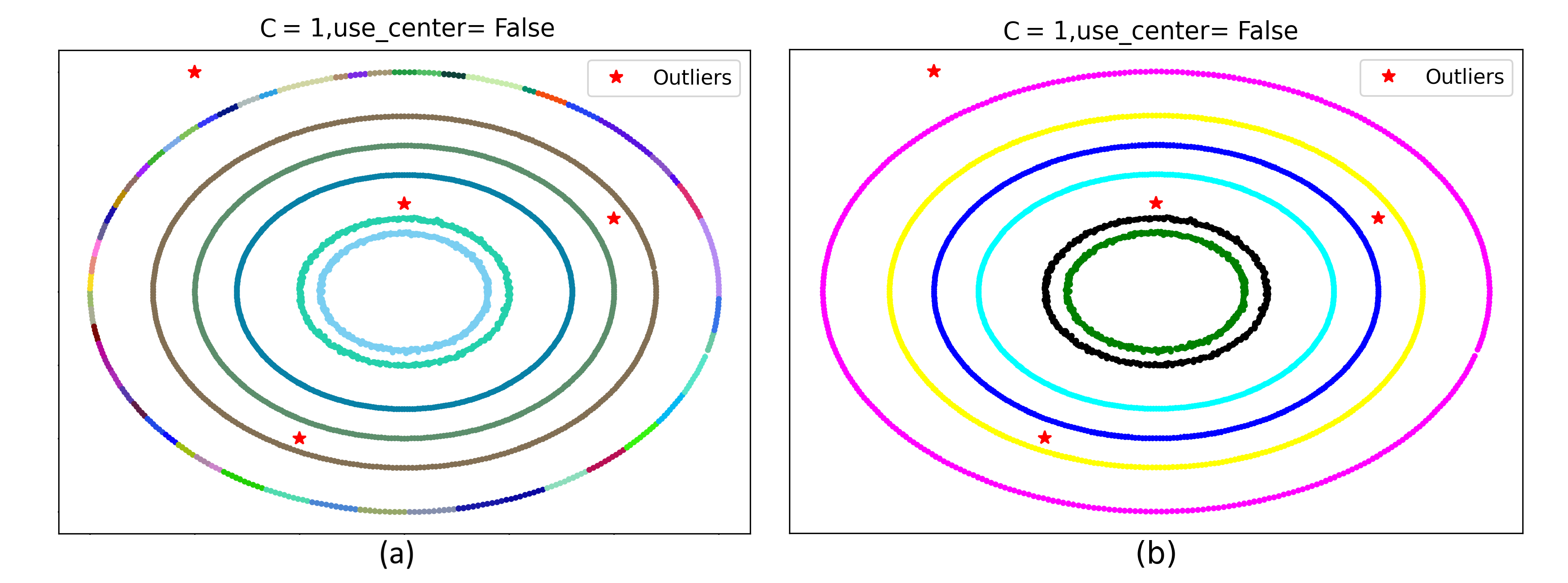}

    \caption{(a) Clustering result without ceiling. (b) Clustering result with ceiling operation (original formulation).}
    \label{fig:appendix}
\end{figure}

This adjustment modifies the resolution of the local neighborhood detection, particularly for datasets with very small or uniform inter-sample distances. Our test showed that this change had a noticeable impact on clustering quality. When the ceiling function was removed, the algorithm failed to clearly distinguish the outer cluster as a whole (Fig. \ref{fig:appendix}(a)), whereas this cluster was well-defined with the original formulation (i.e., with the ceiling function; Fig. \ref{fig:appendix}(b)). 

\FloatBarrier

Thus, the inclusion of \texttt{ceiling} in the computation of \( T \) seems crucial for maintaining consistent clustering performance in datasets with nearly uniform spacing. This ceiling ensures a slightly larger threshold, promoting stability and better separation between dense regions. Removing it leads to underestimation of the neighborhood range, which degrades cluster cohesion.

\section*{Appendix 2} 

\textbf{Formal definition of \(T\) as a density boundary}\\

A standard proxy for local sample density $\rho_i$ at \(p^i\) is inversely related to the $n$-th power of its nearest-neighbor distance:
\begin{equation}
\rho_i \;\propto\; \frac{1}{r_i^n}
\label{eq:density-proxy}
\end{equation}
Indeed, under a locally homogeneous sampling assumption, the expected number of neighbors ($w_i$) within any radius \(\varepsilon>0\) is
\begin{equation}
\mathbb{E}[w_i \mid \rho_i,\varepsilon] \;\approx\; \rho_i\, V_n\, \varepsilon^n \;\approx\; \Big(\tfrac{\varepsilon}{r_i}\Big)^{\!n},
\label{eq:nn-expectation}
\end{equation}
where \(V_n=\pi^{n/2}/\Gamma(\tfrac{n}{2}+1)\) is the volume of the $n$-dimensional unit ball. Two consequences follow immediately:
\begin{align}
\mathbb{E}[w_i \mid r_i=L,\varepsilon=L] \;\approx\; 1, 
\qquad
\mathbb{E}[w_i \mid r_i=\sigma_{\text{opt}},\varepsilon=L] \;\approx\; \Big(\tfrac{L}{\sigma_{\text{opt}}}\Big)^{\!n}\!\!.
\label{eq:contrast}
\end{align}

Thus, choosing the radius \(\varepsilon = L\) to define a ball-shaped neighborhood around a point $p^i$ yields a sharp contrast: extremal sparse points (with \(r_i\approx L\)) have on average \(\lesssim 1\) neighbor inside \(\varepsilon\), while core points (with \(r_i\approx \sigma_{\text{opt}}\)) have many neighbors, growing like \((L/\sigma_{\text{opt}})^n\). This motivates using \(\varepsilon\approx L\) as a \emph{density boundary discriminator}.

To make the procedure scale-free and unit-invariant, we express \(\varepsilon\) in units of \(\sigma_{\text{opt}}\) resulting in a threshold proportional to $\frac{L}{\sigma_{opt}}$. When we defined the threshold $T$ in (\ref{eq:Equation6}), we added a factor $1/2$ and took the ceiling of that value. This additional transformation ensures that applying the threshold T always keep at least all points that are closer than $2\sigma_{opt}$. This limit case is obtained when $L \in \left[\sigma_{opt}, 2\sigma_{opt}\right]$. In all other cases (i.e., $L > 2\sigma_{opt}$), the threshold is more generous, keeping points that are closer than $2\alpha\sigma_{opt}$ when  $L \in \left]2\alpha\sigma_{opt}, 2(\alpha+1)\sigma_{opt}\right]$ for $\alpha \in \{1, 2, 3, 4, ...\}$.

This global threshold $T$ is then used to select a set of close neighbors in the normalized space
\begin{equation}
n_i \;=\; \arg\!\left( \frac{1}{2}\left(\frac{D_i}{\sigma_{\text{opt}}}\right) < T \right)
\;\;\Longleftrightarrow\;\;
n_i \;=\; \arg\!\left( d_{ij} < \varepsilon_T \right),
\qquad \varepsilon_T := 2T\,\sigma_{\text{opt}}.
\label{eq:neighbors-appendix}
\end{equation}
With \eqref{eq:Equation6}, \(\varepsilon_T \approx L\) (exactly \(L\) without the ceiling), so the neighbor test in \eqref{eq:Equation7} implements the principled choice \(\varepsilon\approx L\) justified above.

\section*{Appendix 3} 

\textbf{Effect of cluster density heterogeneity and inter-cluster margin}
\label{sec:margin}

\begin{figure}[!htp]
    \centering
    \includegraphics[width=0.6\linewidth]{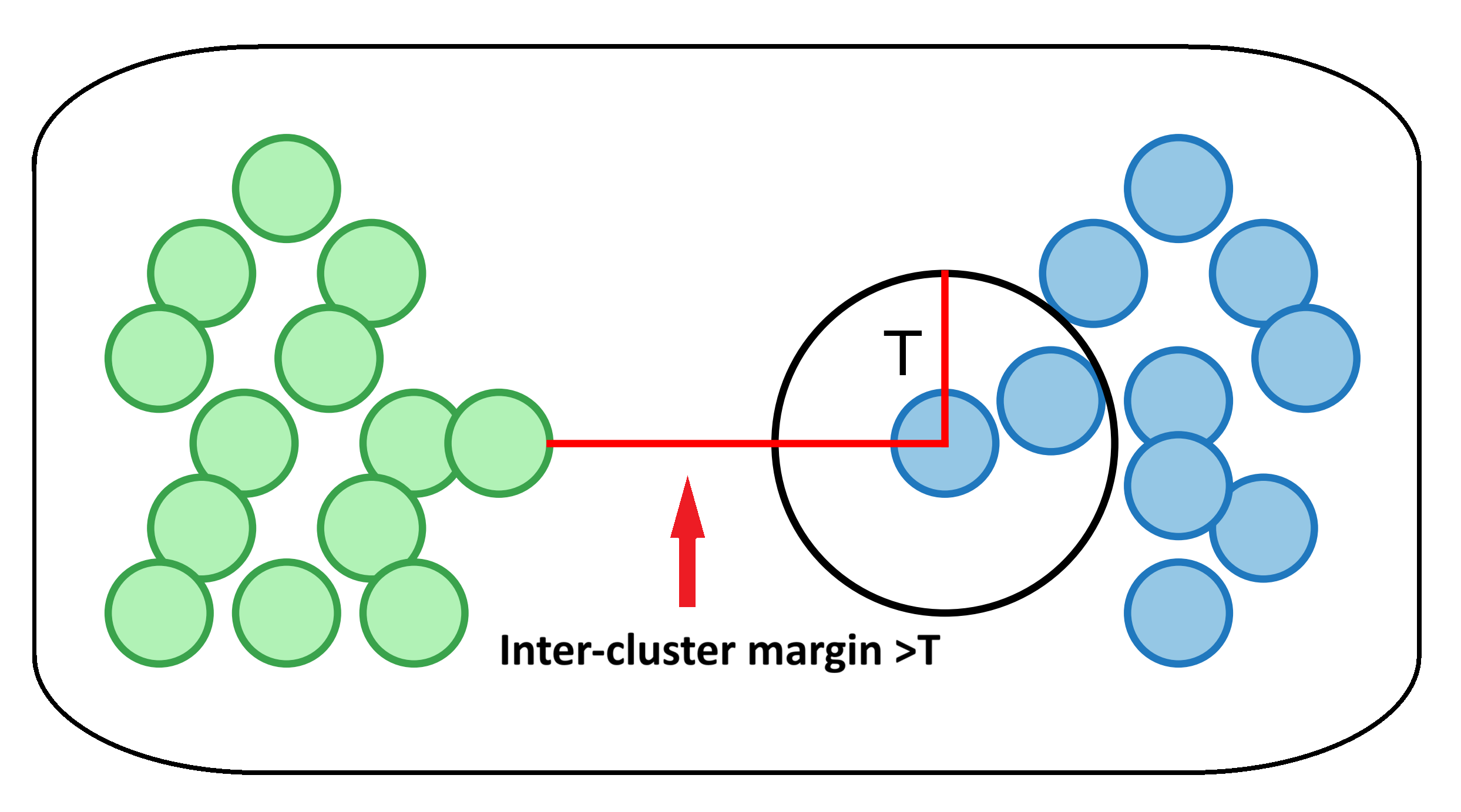}
    \caption{Illustration of some of the concepts involved by SACA, including $T$ and the inter-cluster margin.}
    \label{fig:Inter-cluster margin}
\end{figure}

\noindent The inter-cluster margin $\delta$ (also known as single linkage) between two clusters \( C_i \) and \( C_j \) is defined as \cite{jain1988algorithms}:

\begin{equation}
\delta(C_i, C_j) = \min_{p \in C_i,\, q \in C_j} \|p - q\|
\label{eq:margin}
\end{equation}

\noindent where \( \|p - q\| \) denotes the Euclidean distance between points \( p \) and \( q \). This concept plays a vital role in the effectiveness of SACA. If the following condition 

\begin{equation}
     \delta(C_i, C_j) > T
    \label{eq:Margin_T}     
\end{equation}

\noindent is satisfied, the algorithm can perform clustering with high precision, eliminating the need for parameter selection and adjustment. In other words, when the distance between clusters exceeds the neighbourhood radius $T$, then points near the edges of one cluster will not detect or connect to points from the other cluster (see Fig. \ref{fig:Inter-cluster margin}). This ensures that clusters remain well-separated, allowing the algorithm to achieve high precision without requiring parameter tuning or adjustment.  However, in datasets with overlapping or closely spaced clusters, the algorithm relies on the parameter $C$ to enforce a margin, which can result in the formation of disconnected core regions, as previously illustrated in Fig. \ref{fig:Figure 2}.

\FloatBarrier
In clustering, the true cluster memberships are unknown in advance. Consequently, the inter-cluster margin cannot be computed directly prior to the clustering process. Nevertheless, we have attempted to empirically validate (\ref{eq:Margin_T}) through a series of experiments on the Flame dataset.

\begin{figure}[!htp]
    \centering
    \includegraphics[width=\linewidth]{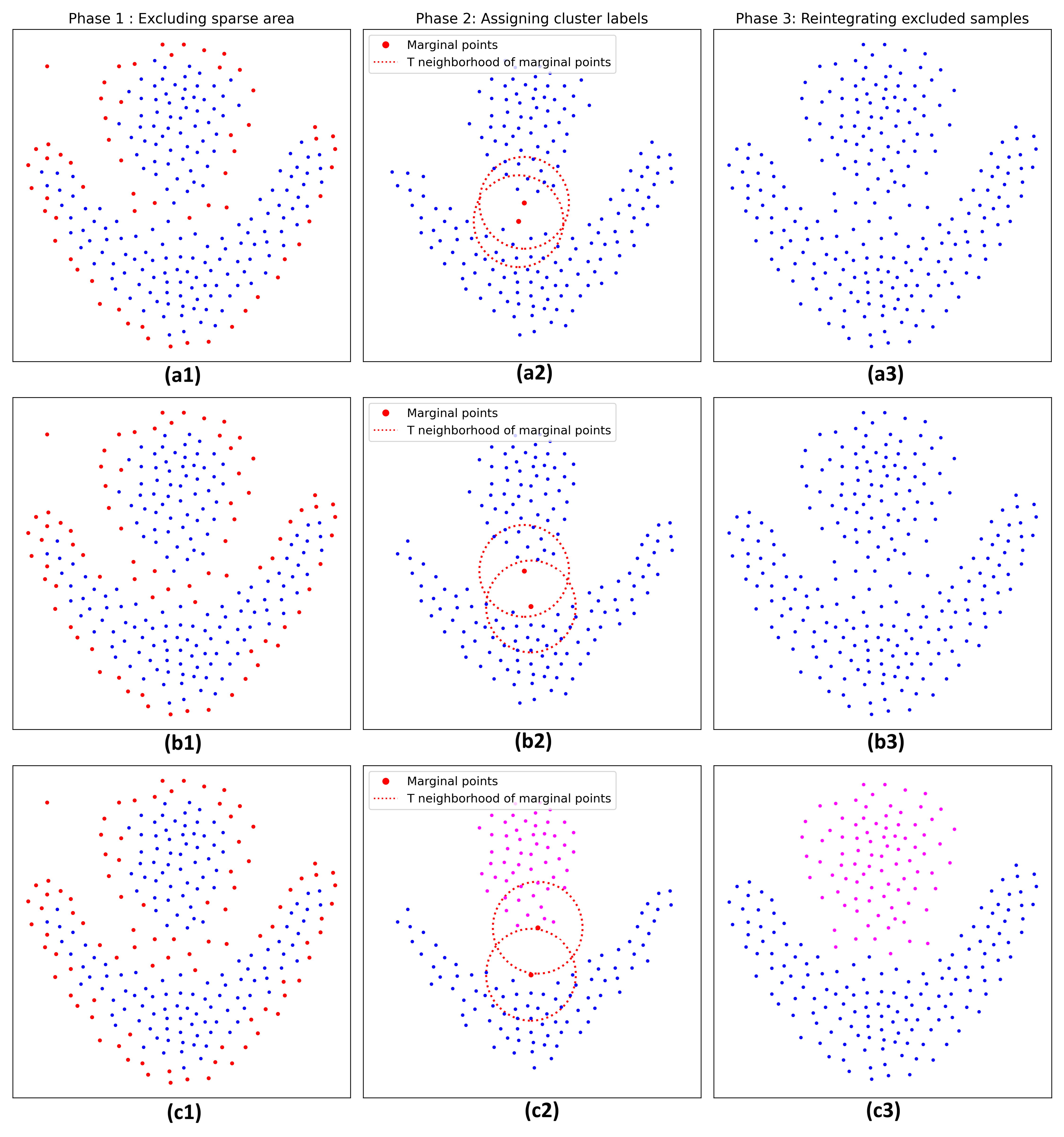}
    \caption{Visual demonstration of (\ref{eq:Margin_T}). Each row of the figure corresponds to a run with specific value for $C$.}
    \label{fig:margin-experiment}
\end{figure}

The computed threshold $T$ for the \textit{Flame} dataset is equal to 2. In the first run with $C=11$, Fig. \ref{fig:margin-experiment}(a2) shows two points indicated by arrows, representing the closest pair of points between two potential clusters (i.e., the margin ), with a Euclidean distance of 0.83. Since $T > \text{margin}$, the clusters are not separable. This occurs because when these points search for their neighbors, they identify each other within radius $T$, resulting in a single cluster.
As the value of $C$ increases, more points are pruned. However, in the second run with $C=12$, the margin with 1.57 still remains smaller than $T$; hence, separation has not yet occurred (Fig.\ref{fig:margin-experiment}(b2)). In the third run, with C=13 and after further pruning, the margin increases to 2.07, exceeding $T=2$. At this stage, since the condition stated in (\ref{eq:Margin_T}) is satisfied, the clusters become successfully separable, as shown in Fig. \ref{fig:margin-experiment}(c2).

\FloatBarrier

\end{document}